# MemGuard-Alpha: Limits of Membership Inference for Detecting and Filtering Memorization-Contaminated Signals in LLM-Based Financial Forecasting

Dip Roy[1*], Rajiv Misra[1], Sanjay Kumar Singh[2], Anisha Roy[3]

[1]*Department of Computer Science and Engineering, Indian Institute of Technology, Patna, India*

[2]*Department of Computer Science, Rajarshi School of Management Technology, Varanasi, India*

[3] *Department of Electronics and Communication Engineering, Jaypee Institute of Information Technology, Noida*

*Corresponding author: dip_25s21res37@iitp.ac.in

**Abstract**

The use of large language models (LLMs) for generating financial forecasts and alpha signals is becoming very popular. However, there is some evidence now suggesting that many LLMs have memorized the historical financial data contained in their training corpora, which leads to overfitting and the production of spurious predictive accuracy that will collapse once the models are moved into an out-of-sample testing environment. This memorization-induced look-ahead bias presents a serious challenge to the validity of using LLMs as part of a quantitative strategy. Membership inference attacks (MIA) have been proposed as a diagnostic for this problem, and remedies such as temporally constrained retraining or input anonymization have been suggested. What has not been established is whether MIA scores are actually informative about memorization in this setting, or whether signal-level filtering built on them improves anything once realistic costs are applied.

We introduce MemGuard-Alpha, comprising the MemGuard Composite Score (MCS), which combines five MIA methods with a temporal proximity feature, and Cross-Model Memorization Disagreement (CMMD), which exploits the natural variation in training cutoff dates across multiple LLMs. We then subject both to an audit that, to our knowledge, has not previously been applied to MIA-based financial memorization detection, and the audit is where the substance of this paper lies. Across seven LLMs (124M to 7B parameters), 50 S&P 100 constituents, 42,800 prompts and 299,600 prompt-model MIA scores spanning January 2019 to June 2024, three findings emerge. First, when in-sample status is defined by a training cutoff, a temporal proximity feature recovers that label perfectly (ROC-AUC 1.000) because it is a monotone transform of the variable that defines it; any composite score containing such a feature therefore reports separation that is arithmetic rather than detection. Second, and more consequentially, the discriminative power of the MIA scores themselves is largely attributable to differences in scale between models rather than to memorization: asked at a fixed date which models had that date in training, raw scores appear highly informative (AUC up to 0.99), but under three independent within-model normalizations discrimination falls to 0.487–0.537. Third, after applying transaction costs symmetrically, no filtering variant improves risk-adjusted performance over the unfiltered ensemble, and none attains significant Fama-French five-factor alpha; simply excluding the weakest model outperforms every contamination-based filter we test. We report these as negative results, together with the failure modes that produced them, and release all artifacts required to reproduce them.

**Keywords:** *Look-ahead bias; Membership inference attack; LLM memorization; Financial alpha signals; Portfolio debiasing; Cross-model disagreement; Quantitative finance*

---

## Note on this version (v2)

---

This version supersedes v1 (March 2026) and reverses its central conclusion. An audit of the original experimental artifacts, prompted by peer review, identified three defects in the portfolio backtest of v1. Transaction costs were applied to the baseline strategies but not to the proposed method, which was scored gross. The 93 evaluation dates lie approximately 21 calendar days apart across a 5.48-year span and were annualised as though consecutive, inflating every Sharpe ratio by roughly a factor of 4.5. The composite score saturates in double precision, so on 56.7% of stock-date pairs the median split admitted every model to the clean group and the proposed filter was identical to the unfiltered baseline. Correcting all three reverses the portfolio result: the Sharpe ratio of 4.11 and the reported 49% improvement in v1 do not survive, and excluding the single weakest model outperforms every contamination-based filter. The evaluation has been extended from 93 to 277 dates. A small number of further quantities reported in v1 could not be reconstructed from the retained artifacts and have been replaced. Readers who have cited v1 should treat its portfolio claims as withdrawn. A corrections record listing every changed quantity, together with a script that regenerates each reported number and prints it beside the v1 value, is included in the replication package.

# 1. Introduction

Financial analysis now uses large language models (LLMs) as a tool of significant power. They also perform better than traditional methods on tasks including directional forecasting [3], earnings predictions [2] and extracting sentiment [1]. As evidence of how rapidly LLMs are being adopted within quantitative finance, there was an increase of 594% between 2023-2025 in LLM-for-finance-related academic publishing (from 36 papers to 250 papers) [4] in leading ML and NLP conferences. Similarly, industry deployment of LLM-based signal generation has been implemented into hedge fund, proprietary trading firm and fintech platform investment pipelines.

The results of all prior studies are threatened by an important theoretical and empirical methodology dilemma. Lopez-Lira et al. [5] showed how GPT-4o is able to recall the exact S&P 500 closing price with less than 1 percent error rate for time frames contained within the training window, while it is significantly worse at doing so for time frames after the training cutoff. As a result, when researchers use historical data to test whether trading signals generated by LLMs perform well, high quality performance metrics may be attributed to the fact that the model has memorized the trading signal outcomes it had seen during the training period, and is therefore not making actual analysis of the market behavior. Benhenda [6] provided empirical support for this by creating a common metric to evaluate LLMs. He demonstrated that the same LLMs provide returns of over 44 percent on their respective in sample periods, however, they experience extreme decline in returns once they have moved into the out of sample period. In particular, the returns from the DeepSeek model were reduced by approximately 22 percentage points past the training cutoff date. Benhenda [6] referred to this phenomenon as the "scaling paradox" because the typical model experiences deterioration in performance as it scales due to the increased influence of prior knowledge that was memorized, whereas the Point-In-Time (PIT) models benefit from scaling due to less reliance on prior knowledge.

The extent of the issue was assessed using a systematic study of all (n = 164) financial LLM research published from 2023 through 2025 on top ML, NLP and AI venues [4] . The study determined that there is no one type of bias (look-ahead, survivorship, narrative, objective, or cost) that has been studied at an incidence rate greater than 28%. Additionally, a practitioner survey of 112 participants was conducted and it was discovered that 74% of practitioners indicated that they do not have available (or have limited access to) ready to use evaluation tools for identifying this type of bias. Furthermore, 50% of the surveyed practitioners stated that they perceive the lack of tools and frameworks to be the largest barrier to mitigating their LLM's bias. The discrepancy between the magnitude of the problem and the availability of solution strategies for resolving the problem is what motivated the current work.

There was additional research using the FINSABER framework, which was presented at KDD 2026, to provide additional support for their findings. The authors of the FINSABER paper created a 20 year backtested pipeline using bias mitigated methods. The main finding of this study is quite clear; as long as the evaluation methodology used is narrow and biased, performance differences in LLM derived alpha will be exaggerated. Once these methodologies are removed through bias mitigation, the performance differences in alpha derived from LLM's vanish. On their evidence the LLM strategies they test do not outperform once these biases are removed, and they attribute the earlier gains to survivorship and look-ahead bias. We report their finding rather than endorse a general claim about market efficiency, on which our own results are silent.

Several approaches to mitigating look-ahead bias have been proposed, each with significant trade-offs. Model retraining approaches, including Time Machine GPT [8], ChronoBERT and ChronoGPT [9], and PiT-Inference models [6], train language models from scratch with strict temporal cutoffs. He et al. [9] use theirs to measure the quantity everyone else assumes: comparing chronologically consistent models against a much larger conventionally trained model on next-day return prediction from financial news, they find comparable Sharpe ratios and conclude that lookahead bias is modest in that setting. This is the most direct estimate of the size of the problem we are aware of, and it bears on how much a filtering method could recover even if it worked perfectly. While effective, this approach is prohibitively expensive at frontier model scales—training a single GPT-4-class model costs millions of dollars. Anonymization approaches remove identifying information: entity-neutering [10] replaces firm names and dates with placeholders, and the BlindTrade framework [11] extends this to multi-agent portfolio construction with anonymized tickers. However, Wu et al. [12] demonstrated that anonymization introduces significant information loss that can be more severe than the look-ahead bias it seeks to address, particularly when numerical and entity information is removed. Inference-time approaches such as divergence decoding [13] modify model behavior by adjusting logits using auxiliary models, but require training model pairs for each unlearning target.

The prior research most directly related to this project is the research by Gao, Jiang, and Yan in [14] which was a statistical test for look ahead bias using membership inference attack (MIA) scores. In their study, Gao et al. introduced Lookahead Propensity (LAP), a Min-K%-based estimate of how likely a model is to have already internalised the outcome for a given firm-date pair, and showed that memorization can amplify what looks like predictive power in LLM-generated forecasts. Most importantly, they demonstrated that memorization is able to operate via a separate mechanism than the model's own internal confidence score. They provided a method to use MIA as an established tool for diagnosing financial memorization; however, Gao et al. did not remove contaminating signal or assess the portfolio level impact.

No prior work builds an end-to-end pipeline that detects memorization-driven signals and filters them before they enter a portfolio, then measures the effect on realised trade returns. We build that pipeline, which requires no retraining, discards no input information, and attaches to any open-weight model as a post-generation layer. It comprises two components: the MemGuard Composite Score (MCS) and Cross-Model Memorization Disagreement (CMMD). Having built it, we report that it does not deliver the portfolio benefit it was designed to deliver, and we identify why.

Machine-learning methods for financial forecasting are a substantial literature in this journal, including graph-convolutional trend prediction [37], multi-source sentiment models for index movement [36] and graph-attention approaches to stock recommendation [38]. The concern this paper addresses applies to all of them once an LLM enters the pipeline. Ludwig et al. [34] give the condition precisely: valid prediction from LLM outputs requires no leakage between the model's training corpus and the researcher's evaluation sample. The problem is best framed as one of measurement. Apparent predictability of next-day returns from a model that has already seen those returns is not predictability in any economically meaningful sense; it is recall presented as inference. The question this paper asks is whether membership inference can tell the two apart at the level of an individual signal, which is what any filtering scheme must assume. Our answer is largely negative, and Section 5 sets out the three reasons. We note here only that the earlier version of this work described MemGuard-Alpha as a measurement instrument capable of separating LLM-derived alpha that survives the efficient market hypothesis from alpha that does not. We withdraw that description. It presupposes that the contamination score measures memorization, which is the proposition under test.

### *1.1 Research Questions*

**RQ1 (Memorization Detection):** Is a MIA-based contamination score able to reliably detect whether an LLM financial prediction was memorized versus created, and is the ability of the MIA-based contamination score to detect memorization affected by different MIA algorithms, LLM families, and the scale used in parameters?

**RQ2 (Impact on Portfolio Performance):** Once transaction costs are applied symmetrically to every strategy, does contamination-based filtering improve out-of-sample portfolio performance relative to the unfiltered ensemble, to threshold filtering, and to the trivial baseline of excluding the weakest model?

**RQ3 (Scale and Temporal Effects):** How does contamination due to memorization differ for different levels of model scale, LLM architecture family, and the most recent year of training data that was used, and are larger models more likely to memorize financial information?

**RQ4 (Robustness):** Are any observed debiasing benefits robust to the choice of contamination threshold, to the partition rule, and to a random-partition control; and what statistical power does a study of this size actually have to answer RQ2?

### *1.2 Contributions*

Our contributions are as follows.

**(i) A tautology in cutoff-based MIA evaluation.** Where in-sample status is defined as $t < c_k$ for a model cutoff $c_k$, any feature monotone in $c_k - t$ recovers the label exactly. Our temporal proximity feature attains ROC-AUC, PR-AUC and TPR at 1% FPR all equal to 1.000. The failure is conditional rather than universal: it arises when a score consumes cutoff-derived features and is then evaluated against cutoff-defined labels. Gao et al. [14] avoid it by construction, since their Lookahead Propensity is computed from answer-position probabilities on a date-only recall query and the cutoff enters only as a placebo window. Ours does not, and any composite that mixes temporal metadata with membership-inference statistics inherits the same defect, which here inflates the reported effect size by roughly an order of magnitude.

**(ii) MIA discriminates between models, not between memorized and unmemorized prompts.** Duan et al. [31] show that membership inference performs close to chance on LLM pretraining data, and Das et al. [32] show that blind attacks which never query the model outperform state-of-the-art membership inference on nine published evaluation datasets, attributing both to distribution shift between the member and non-member samples. We ask whether the same holds in a deployed financial forecasting setting, using a test their designs do not admit. Holding the date fixed removes any temporal confound and poses the question CMMD actually depends on (Fig. 5): can MIA scores identify which models had that date in training? Raw scores suggest they can, with AUC between 0.64 and 0.99. After normalizing within model by z-score, by rank, or by demeaning on model and prompt type, discrimination falls to 0.487-0.537 across all five methods and 33,300 label-varying groups. What the raw scores were picking up is model identity. This is the mechanism Duan et al. and Das et al. identify, arising here from a source their benchmark settings do not contain: systematic differences in score scale between models of different families and sizes evaluated on the same prompt. Within-model evaluation tells the same story: five of seven models fall below chance, and the sole apparent exception dissolves under a sharper test. Restricting to windows around its cutoff, where memorization should be easiest to detect and slow drift hardest, its AUC falls from 0.93 to 0.38.

**(iii) A failure mode in median-split ensemble filtering.** With a large weight on the temporal feature, the logistic link saturates in double precision and 35% of composite scores equal 1.0 exactly. The per-date median then coincides with that saturated value, every model falls on the clean side of the split, and the filter quietly reduces to the unfiltered ensemble on 56.7% of stock-date pairs. The correspondence between the two conditions is exact across all 13,850 groups. Any contamination filter that splits a bounded probability at a sample median is exposed to this.

**(iv) A portfolio-level null, reported with its precision.** Applying transaction costs symmetrically and annualizing over the realized sampling calendar, no filtering variant beats the unfiltered ensemble on a risk-adjusted basis, and none attains significant five-factor alpha. Excluding the single weakest model outperforms every contamination-based filter we test. A minimum-detectable-effect analysis indicates the design would need roughly 862 informative trading days to resolve the observed effect at 80% power, against the 277 available, so we report a bounded null rather than an underpowered positive. This is the concern Bailey et al. [35] formalise for backtests generally: with enough configurations examined, a favourable Sharpe ratio is available by selection alone, and the number of trials must be reported alongside the result. We examined several partition rules and cost conventions before arriving at the corrected figures, which is precisely the situation their analysis describes. Empirically this rests on the largest evaluation of MIA-based memorization detection in finance we are aware of: 7 models spanning 124M to 7B parameters, 50 stocks, 42,800 prompts, 299,600 MIA scores and 5.5 years of daily data.

## 2. Related Work

### *2.1 Look-Ahead Bias in Financial LLMs*

The memorization problem of financial LLMs has been studied for the first time systematically by Glasserman and Lin [15] which have evaluated the look-ahead bias in GPT-generated sentiment analysis. Lopez-Lira et al. [5] were able to provide the most complete evidences until now, demonstrating that GPT-4o can remember with great accuracy (and at times almost perfectly) the closing price of the S & P 500, the date of Wall Street Journal headlines and the level of the stock indexes within the timeframe of his training. They also proved theoretically that, if models memorize the results, the capacity of the model to forecast is non identified — it is impossible to know whether the predictions are due to the knowledge of the results or to the memory of the results. Levy [16] extended this study to numerical reasoning problems. The Federal Reserve's study [17] demonstrated that, in terms of macroeconomic forecasting, they found that the LLMs had a fuzzy temporal awareness — they can approximately remember the dates of the economic calendar, but sometimes they miss them by several days.

Lee et al. [18] uncovered a variety of other biases; they found that LLMs prefer larger cap companies, contrarian investing strategies and that they will show confirmation bias in their analysis. Cao et al. [19] document a "foreign bias" that runs opposite to the classic home bias: the US-trained ChatGPT is systematically more optimistic about Chinese firms than the China-trained DeepSeek, and significantly less accurate. The mechanism they identify is information availability. The bias is strongest where US media coverage of Chinese firms is thin, attenuates for cross-listed firms, and disappears entirely once Chinese news is supplied in the prompt. This is direct evidence that models trained in different information environments produce systematically divergent signals on the same firm-date, which is the same between-model heterogeneity that Section 5.1 shows dominates the membership-inference statistics. These results further reinforce the idea that LLM's problems with bias in finance, are multi-dimensional and need to be mitigated through systematic means.

The Efficient Market Hypothesis (EMH) provides a key conceptual basis to understand why memorizing past values poses such a significant problem for long-term viability of trading systems based on Large Language Models (LLMs). The implication of the semi-strong version of the EMH is that all publicly available information is reflected in current asset prices; therefore, if a large language model appears to have some predictive ability (i.e., has alpha), it can be inferred that it must have better information processing than other models or be obtaining illegal access to future information. If a model has "memorized" past values, then it will appear to have perfect hindsight while pretending to provide foresight. The violation of the assumptions about the information sets that need to be present when running a valid backtest is a direct consequence of memorization. The FINSABER framework [7] formally articulated this point by showing that when controlling for survivorship bias and look-ahead bias, much of the alpha obtained from LLMs vanishes. The disappearance of that alpha after removing methodological biases is completely consistent with the predictions of the EMH that there should be no risk adjusted excess return persistence once methodological biases are accounted for. Therefore, it is essential to develop systematic methods to identify and remove memorization contaminated signals.

Beyond training model memory for individual items, the interaction between several forms of bias also gives rise to compounding effects that are difficult to disaggregate. Kong et al. [4] found five distinct categories of bias in financial applications for LLM — look-ahead, survivorship, narrative, objective, and cost bias. They found that these biases commonly co-occur within the same study. Their practitioner survey found that 74 percent of respondents considered existing tools to be insufficient for detecting bias. This highlights a critical gap between the advancement of deployment of LLM in finance and the availability of infrastructure for validation. Our MemGuard-Alpha framework directly addresses this gap by providing an automated, computational efficiency pipeline for one of the most important types of bias: look-ahead bias through training data memory.

In addition to this recent body of research on contaminated data, there has been an increasing amount of literature to provide a larger scope. In fact, Magar and Schwartz [26], demonstrated that memorization can be utilized to artificially inflate benchmark results, which provided a framework for demonstrating how training data leakage impacts model evaluations across different areas. Sarkar and Vafa [27] showed empirical evidence that indicates that pre-trained language models have lookahead bias when they are applied to financial tasks. Lastly, Yan and Tang [28] developed DatedGPT (a time aware pre-training technique) as a way to prevent temporal information from leaking into web scale language models.

### *2.2 Membership Inference Attacks for LLMs*

Inference attacks based on membership aim to determine if an individual piece of information was included in a model's training set. The work of Carlini et al. [20, 21] demonstrated both empirically and theoretically the mechanisms through which memorization occurs in language models. Their findings indicated that larger models tend to memorize more training data than smaller ones and that memorization is not evenly distributed throughout a model's training corpus. Instead, it tends to occur within sequences that are infrequently seen. Shi et al. [22] proposed Min-K% Prob as their method of determining memorization in language models during ICLR 2024. This is done by averaging the log-probabilities of the K% most difficult to predict tokens in each sequence. The assumption here is that memorized sequences will consist of uniformly high probability values for each token and that those sequences that were not memorized will have "surprising" tokens. By identifying the K% most difficult to predict tokens in each sequence and then calculating the average log-probability of these tokens, the authors argue that they amplify the memorization signal. Zhang et al. [23] built upon this work by developing per-token calibrated versions of Min-K%, referred to as Min-K++%.

We can use a Reference Model Approach [21] to compare the loss of your Target Model versus a Reference Model's Loss. This will help you isolate any Model-Specific Memorization in your model from any General Language Competence your model has. A systematisation-of-knowledge study [24] reviews methodological practice in this literature and finds that most MIA evaluations rely on post-hoc data collection, where distribution shift between members and non-members inflates apparent attack success. Within this journal the method has been studied chiefly from the privacy and defence side, for instance the RM Learning defence of Zhang et al. [39], rather than from the standpoint of whether a membership signal measures memorization in a deployed prediction task. We are able to mitigate these issues because we use Temporal Cutoffs as Ground Truth for each Model.

Gao et al. [14] were the first to bring membership inference to bear on financial forecasting. Their Lookahead Propensity (LAP) statistic estimates, from a date-only recall query, how likely it is that a model has already internalised the outcome for a given firm-date pair, and they show that forecast accuracy correlates with LAP precisely where lookahead bias is present. Their contribution is diagnostic: LAP tells you that contamination exists, but they do not filter contaminated signals at the level of the individual forecast, nor do they measure what contamination costs at the portfolio level.

Methodologically, MIA has moved from shadow-model training, which requires fitting many surrogate models to imitate the target, toward reference-free or single-reference tests needing only the target's output probabilities. That shift is what makes MIA affordable in a financial setting. Meeus et al. [24] survey these methods and stress that the choice of member and non-member distributions can inflate reported attack success. Three results sharpen this into a direct challenge to the enterprise. Duan et al. [31] evaluate membership inference across model scales and find performance close to chance on pretraining data, attributing earlier positive results to temporal distribution shift between members and non-members. Maini et al. [33] reach the same conclusion from a different direction and argue that inference is tractable at the dataset level but not the sequence level. Das et al. [32] make the sharpest version of the argument: on nine published evaluation datasets, blind attacks that distinguish the member and non-member distributions without querying any model outperform state-of-the-art membership inference, so those evaluations measure the sampling procedure rather than the model. Our results are consistent with all three and extend them in two ways. First, we identify a failure that is distinct from distribution shift: where the label is defined by a training cutoff, a feature computed from that cutoff recovers it exactly, which is a property of the label definition rather than of the sample. Second, we show the confound persists in a setting these papers do not examine, where members and non-members are drawn from the same generative process and differ only in date. We adopt temporal cutoffs as ground truth partly to avoid that distributional confound; Section 5.1 shows this substitutes one problem for another.

MIA has many aspects when applied to financial documents. Memorization can occur at various levels of granularity. At one extreme, the model may be memorizing specific data points (i.e. particular closing prices for specific dates). In another aspect, a model may be memorizing an overall trend or narrative (i.e. technology stocks increased in value during 2021). On the other end, the model may learn about persistent statistical relations (i.e. that increasing interest rates will generally have a negative impact upon growth stocks) which would represent "genuine" financial knowledge as opposed to "memorization." Each of our three methods of MIA are designed to detect memorization at different levels of granularity. The loss-based method will detect broad familiarity. The Min-K% method will target memorization related to specific entities by detecting low probability words. The reference model will isolate the knowledge contained within the model versus the knowledge contained in language itself.

## 2.3 Bias Mitigation Approaches

entity-neutering [10], and BlindTrade [11]. However, Wu et al. [12] demonstrated that anonymization reduces the signal quality of the output (the quality of the output is compromised), and in many cases the information lost during anonymization will exceed the information lost through the bias that it removed. A parameter-level variant has since appeared in the same line of work: Gao et al. [29] fine-tune with LoRA on instruction data built from rational benchmark forecasts, which corrects extrapolation bias rather than memorization but places the intervention inside the model. Didisheim et al. [30] document the same memorization phenomenon in financial analysis tasks. Our technique sits in a fourth category: inference-time post-generation signal filtering. This new category of techniques scores the LLM's output for contamination and filters it prior to integrating the output into the portfolio of generated texts; this new category of techniques requires no modifications to the LLM itself.

Each category of mitigation has its own trade-off. Theoretical prevention using retraining provides the greatest potential to prevent look ahead behavior since if a model has been trained on no post-cutoff data, then the model can't produce look ahead behavior for the same time frame. However, the retraining of models of large sizes increases both the cost of computing the model and the number of times the model will need to be updated. If you are running a hedge fund where signals are being produced and evaluated every day, there would be economic feasibility issues with retraining a frontier language learning model for every evaluation day. Daily retraining of a frontier model is not economically feasible for a signal pipeline of this kind. A second approach to mitigation, called anonymizing, does provide a much cheaper alternative, however; it creates a significant conflict. The very information that allows the model to remember (entity names, specific dates, etc.) is likely the very information that is used in analyzing financial markets. Wu et al. [12] have shown empirically that the information lost by anonymizing data (transcripts of company earnings calls), in most cases, is greater than the bias removed. Inference-time approaches like Divergence Decoding [13], create an alternative that preserves all input information, but modifies the output distribution of the model. However, these inference-time methods require creating a new pair of models for each target domain and time boundary.

Our post-generation filter is an alternative to other approaches to mitigate the risks associated with memorization, since our method doesn't add any computational overhead or slow down the inference pipeline that produces the output. MIA scoring happens in a second pass and could also be run in parallel over different models and prompts. In addition, we are changing the way people think about preventing memorization (which may inherently occur at scale when a large amount of data is used for training), to manage how memorization affects portfolio decision-making. The difference in philosophy, from trying to avoid a problem to trying to manage its impact, fits well into how many financial professionals handle various forms of noise and bias in their quantitative work, including estimating costs of transacting, managing exposures to factors and calibrating risk models. While a risk model does not prevent volatility, but rather helps manage the impact of volatility on constructing portfolios; similarly, MemGuard-Alpha does not prevent memorization, but rather helps manage the negative impact of memorization on signal quality.

## 2.4 Positioning of This Work

Table 1 positions MemGuard-Alpha against prior work across six dimensions. MemGuard-Alpha occupies a position no prior method does, combining detection with signal-level filtering, requiring no model modification, preserving full information content, measuring portfolio-level impact, and operating across models with per-model temporal awareness. The table describes design properties, not demonstrated benefits; Section 5 reports that the portfolio-level benefit does not materialise.

**Table 1.** *Positioning of MemGuard-Alpha against existing approaches.*

| Approach | Type | Model Mod. | Info Loss | Portfolio Test | Multi-Model |
|---|---|---|---|---|---|
| ChronoBERT/GPT [9] | Prevention | Full retrain | None | No | No |
| PiT-Inference [6] | Prevention | Full retrain | None | Yes | No |
| Entity-Neutering [10] | Prevention | None | High | Partial | No |
| BlindTrade [11] | Prevention | None | High | Yes | Yes |
| Divergence Decoding [13] | Mitigation | Aux. models | None | Partial | No |
| LAP Test [14] | Detection | None | None | No | No |
| **MemGuard-Alpha (Ours)** | **Det.+Filter** | **None** | **None** | **Yes** | **Yes** |

# 3. Methodology

## 3.1 Problem Formulation

Let $M = \{m_1, \ldots, m_K\}$ be a set of K autoregressive language models, each with a known training data cutoff date $c_k$. For a financial prompt x associated with ticker s and date t, model $m_k$ generates an alpha signal $\alpha_k(x) \in \{-1, 0, +1\}$ (bearish, neutral, bullish) with confidence $\gamma_k(x) \in [0, 1]$. The central challenge is that for dates $t < c_k$ (within the training window), $\alpha_k(x)$ may reflect memorization of the outcome rather than genuine reasoning.

We define a contamination scoring function $\phi(x, m_k)$ quantifying the likelihood that $m_k$'s response to x is memorization-driven. MemGuard-Alpha operates in three stages: (1) **Score:** compute $\phi(x, m_k)$ using MIA methods; (2) **Partition:** split model predictions into clean and tainted groups; (3) **Trade:** construct the portfolio using only the clean consensus signal.

## 3.2 Data Sources

We are utilizing three data sets. First, we are collecting daily OHLCV price information on 50 S&P 100 constituents from Yahoo Finance, from January 2019 through June 2024 (approximately 1,375 trading days each stock) in order to cover several different market regimes. The time span includes: the expansion prior to the pandemic (2019); the COVID-19 crash and the subsequent recovery (March – December 2020); the post-stimulus bull market (2021); the Federal Reserve tightening cycle and the bear market (2022); and the current AI-based tech rally (2023 – 2024). Second, we will utilize the Financial PhraseBank corpus [25], with 3,100 human-annotated financial news sentences from financial news prior to 2016, which predates every model's cutoff and therefore serves as a pre-cutoff reference corpus. We described it in the submitted version as a guaranteed-memorized control. That was too strong: publication before a cutoff does not establish that specific sentences entered any particular pretraining corpus, and Section 5.1 reports a test that bears this out directly. Third, we will collect out-of-sample control prompts from the same templates that were used for all model's training windows (April - June 2024), which provide a control group outside of the sample space.

## 3.3 Model Selection and Per-Model Temporal Cutoffs

We evaluate seven models spanning four architectural families. Table 2 summarizes the lineup. The per-model cutoff design is fundamental to CMMD: the same prompt may be in-sample for one model but out-of-sample for another.

**Table 2.** *Model lineup with training cutoffs and IS/OOS prompt partitions.*

| Model | Parameters | Family | Cutoff | IS Prompts | OOS Prompts |
|---|---|---|---|---|---|
| GPT-2 | 124M | GPT-2 | Oct 2019 | 6,200 | 36,600 |
| GPT-2 Medium | 355M | GPT-2 | Oct 2019 | 6,200 | 36,600 |
| GPT-2 Large | 774M | GPT-2 | Oct 2019 | 6,200 | 36,600 |
| TinyLlama 1.1B Chat | 1.1B | LLaMA | Sep 2023 | 35,750 | 7,050 |
| Phi-2 | 2.7B | Phi | Oct 2023 | 36,350 | 6,450 |
| Qwen2.5 3B Instruct | 3B | Qwen | Mar 2024 | 39,500 | 3,300 |
| Qwen2.5 7B Instruct | 7B | Qwen | Mar 2024 | 39,500 | 3,300 |

The GPT-2 family (cutoff Oct 2019) has only 14.5% in-sample prompts. TinyLlama and Phi-2 (cutoffs Sep–Oct 2023) have ~83% coverage. Qwen models (cutoff Mar 2024) have 92% coverage. This gradient is essential for RQ3.

Three criteria governed model selection. First, architectural diversity: the seven models span four families (GPT-2, LLaMA, Phi, Qwen). We originally described this as ensuring that observed differences would not be attributable to any single architecture. Section 5.1 shows the opposite is closer to the truth, since between-model differences turn out to dominate the MIA statistics, and we retain the diverse lineup because it makes that confound visible rather than because it eliminates it. Second, temporal diversity: three distinct cutoffs (October 2019, September to October 2023, March 2024) allow comparison within a family at fixed cutoff (the three GPT-2 models) and across families at similar scale but different cutoff (GPT-2 Large against TinyLlama). Third, accessibility: all seven are open-weight, which is required because Min-K%, Min-K%++ and the reference method need per-token log-probabilities that commercial APIs do not expose. Section 6.5 reports what remains achievable when they are unavailable.

All seven models, including the 7B Qwen model, were loaded at full precision; no quantization was applied at any stage. An earlier version of our pipeline contained a 4-bit loading path for models above 7B parameters, but that path was removed before the production run and never executed. We note this explicitly because the distinction matters for the reproducibility of the MIA scores, which are sensitive to numerical precision.

### 3.4 MIA Scoring Engine

We implement five established MIA methods, each capturing a different aspect of the memorization signal. For a text sequence $x = (x_1, \ldots, x_n)$ and model m with vocabulary V:

**Loss.** Average negative log-likelihood:

$$L(x, m) = -(1/n) \sum_i \log p(x_i \mid x_1, \ldots, x_{i-1}; m) \qquad (1)$$

Lower loss indicates higher familiarity, suggesting the text was encountered during training. This is the simplest and most computationally efficient MIA method.

**Min-K% Prob [22].** Rather than averaging over all tokens, Min-K% focuses on the K% (K=20) tokens with the lowest log-probabilities. The intuition is that memorized text will have uniformly high token probabilities, while non-memorized text will have some tokens that the model finds surprising. By focusing on these "hardest" tokens, Min-K% amplifies the memorization signal, particularly for entity-specific tokens (exact prices, specific dates) that are disproportionately affected by memorization.

**Min-K%++ [23].** Extends Min-K% by applying per-token calibration to each token's position. Each token position's log-probability is then normalized using a z-score against the overall mean and standard deviation of log-probabilities across all tokens in the vocabulary at that specific position. The normalization captures differences in vocabulary-level difficulty (a token appearing as having very low probability may simply represent how rare the token is). Thus, Min-K%++ is able to separate model-specific familiarity from inherent token difficulty.

**Zlib Ratio.** Ratio of model loss to zlib compression entropy: $L(x, m) / H_{zlib}(x)$. Zlib compression provides a model-free estimate of text complexity. The ratio normalizes for text structure, distinguishing genuine memorization (low loss on complex text) from easy prediction (low loss on simple text). (2)

**Reference [21].** Ratio of target model loss to reference model (GPT-2 base) loss: $L(x, m_{target}) / L(x, m_{ref})$. Values substantially below 1.0 indicate that the target model is more familiar with the text than a generic language model, suggesting model-specific memorization. (3)

### 3.5 MemGuard Composite Score (MCS)

No single MIA method dominates across all models (see Section 5.1). MCS combines all five with temporal proximity via logistic regression:

$$MCS(x, m_k) = \sigma(w^T [\phi_1, \ldots, \phi_5, \tau(t, c_k)] + b) \quad (4)$$

where $\tau(t, c_k) = \min(\max((c_k - t)/1825, -1), 1)$ normalizes temporal distance to $[-1, 1]$, and w, b are learned on IS/OOS labels. Two properties of this construction matter for what follows, and we state them here rather than in the results because they follow from the definition rather than from any experiment. First, $\tau > 0$ holds exactly when $t < ck$, which is the definition of in-sample status, so τ separates the two classes perfectly by construction and any weight fitted to it will be large. Second, because τ is bounded and its fitted weight is large, the logistic link saturates: with a fitted weight of 62.19 on a standardized feature, σ(·) returns exactly 1.0 in double precision over much of the domain. MCS is therefore not a calibrated probability across its full range, and Sections 5.1 and 5.2 quantify what that costs.

### *3.6 Cross-Model Memorization Disagreement (CMMD)*

CMMD is the central algorithmic contribution. For each (ticker s, date t), we observe predictions from K models with varying contamination. Algorithm 1 describes the procedure:

**Algorithm 1: CMMD**

```
for each model mk, compute MCSk(s, t)
med ← median({MCSk})
C(s,t) ← {k : MCSk ≤ med}                              ▷ Clean set
T(s,t) ← {k : MCSk > med}                              ▷ Tainted set
αCMMD(s,t) ← mean({αk : k ∈ C})                        ▷ Trading signal
δ(s,t) ← meanT(αk) − meanC(αk)                         ▷ Disagreement
```

A large positive δ indicates tainted models are more bullish than clean models, suggesting memorization-driven optimism. The design intuition behind CMMD is that differing training cutoffs create variation in memorization status that is not chosen with reference to any particular stock or date. Section 5.2 examines how much of that variation the composite score actually recovers, and the answer bears directly on whether the partition measures memorization or something else.

The median split has a variance rationale. For a stock-date pair with K models ordered by contamination score, the disagreement δ(s, t) estimates any systematic difference between the two groups, and its variance falls with the harmonic mean of the group sizes. An equal split maximises that harmonic mean, so among binary partitions the median is the minimum-variance choice, in the same sense that balanced arms minimise variance in experimental design. This argument is correct as far as it goes and it is why we chose the median. It assumes the score is on a scale where a median is meaningful, and Section 5.1 shows that assumption fails here: once the logistic link saturates, the median coincides with the saturated value and the split is not balanced but empty on one side.

An alternative method to the median split is the use of a "hard" threshold (i.e., classifying all models with an MCS value greater than 0.5 as "tainted"). However, there are two disadvantages to the use of fixed thresholds. First, MCS values are not perfectly calibrated across stocks and dates. That is, the probability distributions of contamination change based upon the number of in-sample models for each given date. Therefore, in early 2019, a date will have only the GPT-2 family of models in-sample and thus have a very different MCS distribution than a date in late 2023 which has six out of seven models in-sample. Thus, the median will adapt automatically to these changes in the probability distributions. Second, a fixed threshold risks degenerate partitions in which every model falls on one side and no discrimination is possible. We report this argument as originally advanced because our own results refute it: the median split is subject to exactly the same failure, and more severely. When the logistic link saturates, the per-date median coincides with the saturated value, the condition MCSk ≤ med admits every model, and the tainted set is empty. This occurs on 56.7% of stock-date pairs (Fig. 3), on which CMMD is identical to the unfiltered ensemble. Adaptivity to the per-date distribution, which motivated the median in the first place, is precisely what propagates the saturation into the partition.

It is tempting to read the median split as a quasi-experimental design, with cutoff dates as an exogenous assignment mechanism and the median as a discontinuity, and an earlier version of this work framed it as a fuzzy regression discontinuity. We now regard that framing as an analogy rather than an identification strategy, and we set it out here only to explain why. A fuzzy RDD requires a running variable that is measured independently of the outcome and a threshold fixed in advance. Neither holds. The running variable, MCS, is fitted on labels derived from the same cutoff dates that supposedly provide the exogenous variation, so it is not independent of the assignment. The threshold is a sample median recomputed for every stock-date pair, so there is no fixed discontinuity around which to test density continuity, and no first stage to report. The intuition that differing cutoffs create useful variation is sound and it is what motivates CMMD; the formal apparatus of regression discontinuity is not available to us, and we do not claim it.

The same objection applies to reading δ(s, t) as a local average treatment effect, and we drop that too. The required exclusion restriction is not merely unverified but implausible, since a model's cutoff is confounded with its release date and hence with architecture, corpus, tokenizer and instruction tuning, all of which may affect signal quality through channels unrelated to memorization. Section 5.2 shows this is not hypothetical. We therefore treat δ(s, t) as a descriptive contrast between two groups of models, and make no claim about market efficiency on its basis.

### 3.7 Portfolio Construction and Evaluation

We construct daily signal-weighted portfolios across 50 stocks with realistic transaction costs. For each trading day t, the position in stock s is proportional to the strategy's signal for that stock. Returns are computed as:

$$R(t) = (1/N) \sum_s \alpha(s, t) \cdot r(s, t+1) - TC \cdot turnover(t) \qquad (5)$$

where TC = 15 basis points (10 bps execution cost + 5 bps slippage) and turnover is measured as the mean absolute change in signal across stocks. This cost model is conservative relative to institutional execution but appropriate for the daily rebalancing frequency of our strategies.

We compare seventeen strategies (Table 4). The core set is Raw Alpha, the unfiltered mean signal across all seven models; Debiased Alpha, which zeroes signals above the median contamination score; and CMMD, the clean-model consensus, reported under both the published median split and the corrected rank split. To these we add ensemble variants (excluding Qwen-7B, the GPT-2 trio alone, and the strictly out-of-sample subset), conventional benchmarks (equal-weight buy-and-hold, 20-day momentum, 5-day reversal, volatility-scaled momentum, random), a market-neutral long-short variant of CMMD, and two walk-forward machine-learning forecasters. Costs of 15 bps are applied identically to every strategy. Significance uses a stationary block bootstrap (10,000 resamples, block length five) on Sharpe differences, reported two-sided throughout.

## 4. Experimental Setup

### 4.1 Computational Infrastructure

All experiments were conducted on a RunPod cloud instance with an NVIDIA GeForce RTX 5090 GPU (32 GB GDDR7, Blackwell architecture, compute capability 12.0). PyTorch 2.10.0 with CUDA 12.8 was used. Total MIA scoring: ~4.5 hours for 299,600 prompt-model pairs. Alpha generation: ~6 hours on an optimized subset. Experiments logged to Weights & Biases with per-model checkpointing. All models were loaded at full precision; the 7B Qwen model fits within the 32 GB device memory without quantization.

### 4.2 Prompt Design

Three prompt types probe different memorization aspects. *Price recall* (e.g., "On January 15, 2021, AAPL stock closed at") tests factual memorization. *Sentiment* (e.g., "AAPL shares rose on January 15, 2021 as investors reacted to") tests contextual memorization. *Forward* (e.g., "Based on AAPL recent performance as of January 15, 2021, analysts expect") tests predictive memorization. Dates were sampled every fifth trading day across 2019–2024.

### 4.3 Alpha Signal Generation

Each model is provided a structured prompt that will ask for both bullish and bearish views in order to make a prediction with confidence. The above balanced format generates 60% bullish, 18% bearish, 22% neutral signals. Parameters used for generating the data: temperature = 0.7; top-p = 0.9; max = 80 new tokens. One template is used per (ticker, date). The submitted version sampled every third MIA-scored date, giving 93 trading days and 5,900 prompts per model; this revision generates the remaining dates, giving 277 days and 13,850 prompts per model, of which 96,950 prompt-model pairs feed the portfolio evaluation.

sides of the issue before making a decision, was intended to minimize the influence of memory based on past sentiment, while maximizing the amount of analytical thought. The submitted version reported that in-sample accuracy substantially exceeded out-of-sample accuracy despite this design. Section 5.5 shows the within-model gaps are between −0.6 and +4.7 percentage points and none is individually distinguishable from zero, so we no longer draw an inference about how memorization enters the reasoning step.

The submitted version generated alpha signals on every third sampled date, giving 93 trading days. For this revision we generated the remaining dates, so signals now exist for all 277 dates on which MIA scores were computed, roughly one every fifteen trading days across the evaluation window and 50 per calendar year. Generation used the identical prompt, decoding parameters and parser; the only procedural change was to batch prompts rather than process them singly. To confirm that batching did not shift the output distribution we regenerated 400 prompts per model from the original 93 dates and compared the bull/bear/neutral distribution against the stored signals for the same stock-date pairs; no model differed at $p < 0.01$ ($\chi^2$), and the bullish rate moved by at most 1.2 percentage points between the original and new dates. Portfolio inference is still limited by sample size, as Section 6.6 quantifies.

### *4.4 Data Preprocessing and Quality Control*

The price data was subject to many of the same quality control processes that are used in all finance research. We performed three basic steps of quality control. Step One: We made corporate action adjustments on the price data using Yahoo Finance's "adjusted" close price, which accounts for stock splits, dividend payments, and other events that affect the number of shares outstanding. Step Two: We excluded the price for any day in our sample where the closing price was missing for any of the 50 tickers we were analyzing. This ensured a complete time series for each of the stocks included in this study. Step Three: We calculated the forward one-day return for each of the stocks. We defined the forward one-day return as $r(s, t + 1) = (P_adj(s, t + 1) / P_adj(s, t)) - 1$, where P_adj is the adjusted closing price. By using the adjusted closing price to calculate the return, it was possible to model what investors actually experienced during their investment horizon. Step Four: We winzorized the daily returns for each of the stocks at both the 0.5th percentile and 99.5th percentile to minimize the impact of extreme outliers on the portfolio level summary statistics. The final dataset included approximately 1375 trading days per stock across the January 2019 through June 2024 evaluation period.

For MIA scoring, text prompts were tokenized using each model’s native tokenizer, ensuring that token-level log-probabilities are computed on the model’s actual vocabulary rather than a universal tokenizer that might introduce distribution shift. Prompts exceeding 512 tokens were truncated to fit within the context windows of smaller models (GPT-2 family), though in practice the financial prompts in our dataset average approximately 45 tokens and never exceed 200 tokens. The five MIA scores (loss, min-k, min-k++, zlib, reference) were computed deterministically with no sampling or temperature, using the model’s greedy log-probability assignments. This ensures full reproducibility: the same prompt processed through the same model will always produce identical MIA scores.

## 5. Results

### *5.1 MIA-Based Memorization Detection (RQ1)*

All 34 valid model-method combinations show significant separation under the Kolmogorov-Smirnov test, though four fail to reach significance on the t-test (Table 3). Figure 1 visualizes the aggregate distributions.

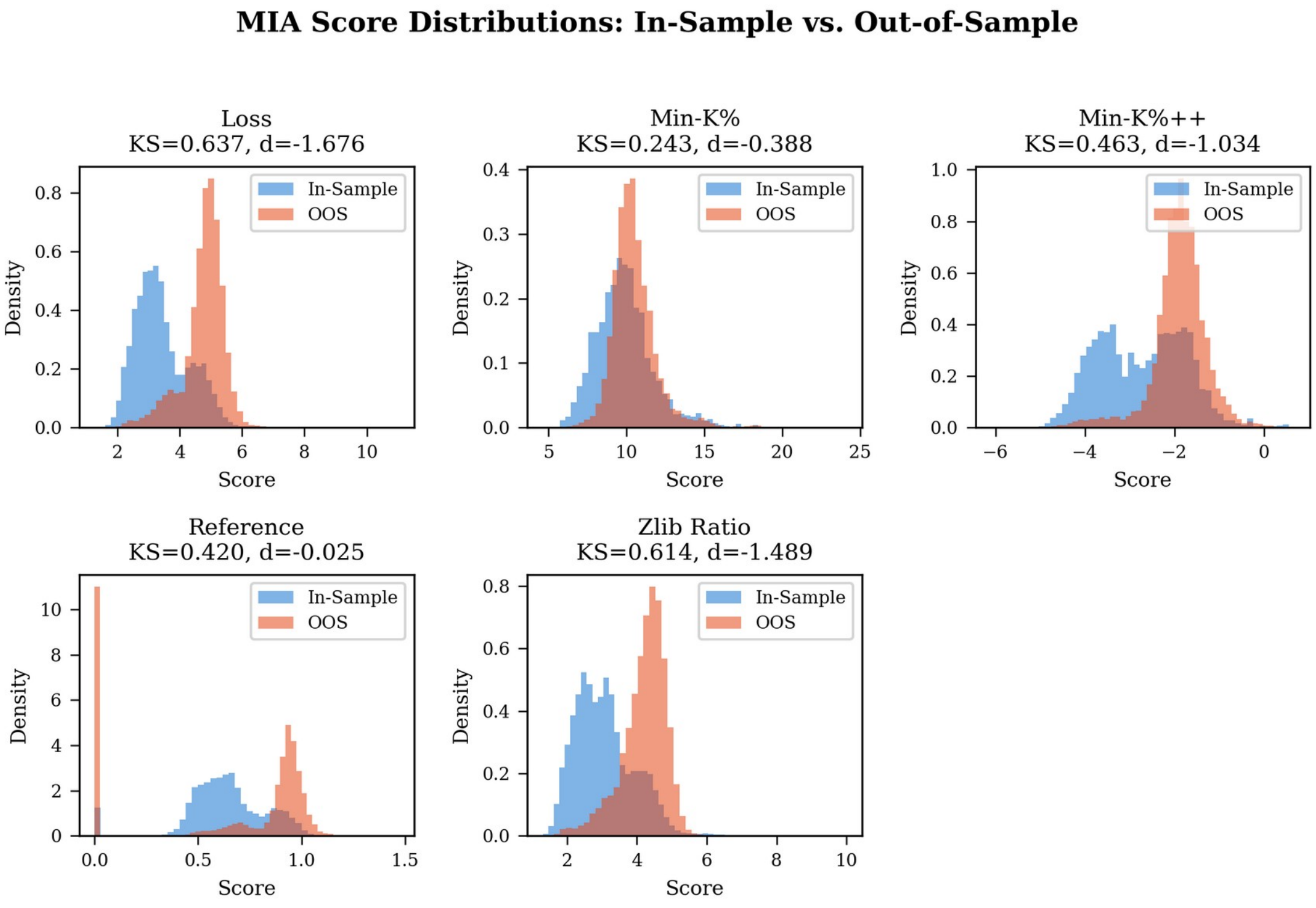

**Fig. 1.** MIA score distributions for in-sample (blue) vs. out-of-sample (red) prompts across five methods. All methods show statistically significant separation (KS $p < 0.001$ for all).

Table 3 reports all 34 valid model-method combinations; the GPT-2 self-reference is excluded because the ratio is trivially 1.0. The strongest separations are TinyLlama on mia_loss ($d = -1.37$), mia_ref (−1.16) and mia_zlib (−1.00).

**Table 3.** *Complete RQ1 results: IS vs. OOS MIA score separation across all model-method combinations. † denotes p > 0.05 (not significant at conventional levels).*

| Model | Method | IS Mean | OOS Mean | Cohen's d | KS p | t-test p |
|---|---|---|---|---|---|---|
| gpt2 | mia_loss | 5.058 | 5.161 | −0.19 | 9.7e−61 | 3.2e−45 |
| gpt2-medium | mia_loss | 4.786 | 4.939 | −0.26 | 1.5e−98 | 4.3e−85 |
| gpt2-large | mia_loss | 4.675 | 4.836 | −0.33 | 1.0e−122 | 5.2e−134 |
| TinyLlama 1.1B | mia_loss | 3.101 | 3.554 | −1.37 | <1e−300 | <1e−300 |
| Qwen2.5-3B | mia_loss | 2.905 | 3.112 | −0.39 | 1.6e−76 | 2.9e−102 |
| Phi-2 | mia_loss | 4.203 | 4.599 | −0.67 | <1e−300 | <1e−300 |
| Qwen2.5-7B | mia_loss | 2.927 | 3.113 | −0.37 | 1.7e−80 | 1.1e−93 |
| gpt2 | mia_min_k | 10.408 | 10.578 | −0.13 | 8.2e−38 | 5.9e−21 |
| gpt2-medium | mia_min_k | 10.301 | 10.652 | −0.22 | 1.8e−156 | 3.5e−63 |
| gpt2-large | mia_min_k | 10.069 | 10.327 | −0.19 | 2.7e−110 | 1.5e−47 |
| TinyLlama 1.1B | mia_min_k | 9.735 | 10.757 | −0.89 | <1e−300 | <1e−300 |
| Qwen2.5-3B | mia_min_k | 9.617 | 9.653 | −0.02 | 3.3e−4 | 0.319† |
| Phi-2 | mia_min_k | 10.372 | 10.833 | −0.22 | 8.3e−103 | 5.4e−55 |
| Qwen2.5-7B | mia_min_k | 9.572 | 9.619 | −0.03 | 8.7e−4 | 0.137† |
| gpt2 | mia_min_k_pp | −1.746 | −1.692 | −0.13 | 2.1e−31 | 5.1e−23 |
| gpt2-medium | mia_min_k_pp | −1.887 | −1.764 | −0.27 | 3.6e−103 | 1.3e−98 |
| gpt2-large | mia_min_k_pp | −1.947 | −1.849 | −0.18 | 5.7e−87 | 7.4e−44 |
| TinyLlama 1.1B | mia_min_k_pp | −1.956 | −1.929 | −0.06 | 9.2e−7 | 1.5e−5 |
| Qwen2.5-3B | mia_min_k_pp | −3.503 | −3.545 | +0.07 | 2.6e−5 | 2.2e−4 |
| Phi-2 | mia_min_k_pp | −2.289 | −2.150 | −0.16 | 3.1e−66 | 3.4e−30 |
| Qwen2.5-7B | mia_min_k_pp | −3.507 | −3.513 | +0.01 | 5.7e−4 | 0.603† |
| gpt2-medium | mia_ref | 0.946 | 0.958 | −0.19 | 2.8e−30 | 1.3e−40 |
| gpt2-large | mia_ref | 0.926 | 0.938 | −0.24 | 3.8e−69 | 1.4e−72 |
| TinyLlama 1.1B | mia_ref | 0.607 | 0.687 | −1.16 | <1e−300 | <1e−300 |
| Qwen2.5-3B | mia_ref | 0.567 | 0.591 | −0.24 | 1.3e−30 | 3.7e−39 |
| Phi-2 | mia_ref | 0.820 | 0.882 | −0.65 | <1e−300 | <1e−300 |
| Qwen2.5-7B | mia_ref | 0.572 | 0.591 | −0.21 | 2.8e−32 | 3.9e−30 |
| gpt2 | mia_zlib | 4.557 | 4.568 | −0.02 | 2.6e−26 | 0.168† |
| gpt2-medium | mia_zlib | 4.311 | 4.371 | −0.10 | 3.6e−75 | 4.5e−14 |
| gpt2-large | mia_zlib | 4.211 | 4.280 | −0.13 | 1.7e−62 | 2.8e−23 |
| TinyLlama 1.1B | mia_zlib | 2.759 | 3.146 | −1.00 | <1e−300 | <1e−300 |
| Qwen2.5-3B | mia_zlib | 2.597 | 2.741 | −0.25 | 1.9e−56 | 2.5e−42 |
| Phi-2 | mia_zlib | 3.741 | 4.067 | −0.53 | 5.0e−252 | <1e−300 |
| Qwen2.5-7B | mia_zlib | 2.615 | 2.741 | −0.23 | 9.8e−44 | 6.0e−36 |

Of the 35 model-method pairs, 34 are valid (the GPT-2 self-reference is excluded), and all 34 KS tests achieve $p < 0.001$. However, four t-test comparisons do not reach conventional significance at $p < 0.05$, marked with † in Table 3: Qwen2.5-3B on mia_min_k ($p = 0.319$), Qwen2.5-7B on mia_min_k ($p = 0.137$), Qwen2.5-7B on mia_min_k_pp ($p = 0.603$), and GPT-2 on mia_zlib ($p = 0.168$). Because Table 3 reports 34 valid tests, we apply a Bonferroni correction: at a family-wise α of 0.05 the per-test threshold is 0.00147. Every KS test clears it by many orders of magnitude, and the four t-tests already flagged as non-significant remain so, making the correction immaterial here. The correlation analysis (Figure 2) reveals mia_loss and mia_zlib are highly correlated ($r = 0.98$) while mia_ref provides independent information ($r < 0.15$).

Two patterns in Table 3 are worth noting. Separation rises monotonically with size within the GPT-2 family ($d = −0.19, −0.26, −0.33$ at 124M, 355M and 774M), consistent with the memorization scaling reported by Carlini et al. [20]: larger models have more capacity for the low-frequency tokens that dominate these prompts, such as specific dates, tickers and prices. Across families the ordering does not follow size. TinyLlama at 1.1B shows the strongest separation of any model ($d = −1.37$), well ahead of GPT-2 Large at 774M, and the natural reading is that recency matters more than scale. We flag that reading as confounded: TinyLlama also differs from GPT-2 in architecture, corpus, tokenizer and instruction tuning, and 83.5% of our prompts fall inside its window against 14.5% for GPT-2. Section 5.2 gives a more direct reason for caution.

Three of the four non-significant t-tests involve Qwen models on Min-K% variants, which suggests that Qwen's tokenisation or attention distributes probability mass in a way that blunts the discriminative power of the lowest-probability tokens while leaving loss- and reference-based methods intact. The fourth, GPT-2 on zlib, most likely reflects the weak memorization signal available to the oldest and smallest model.

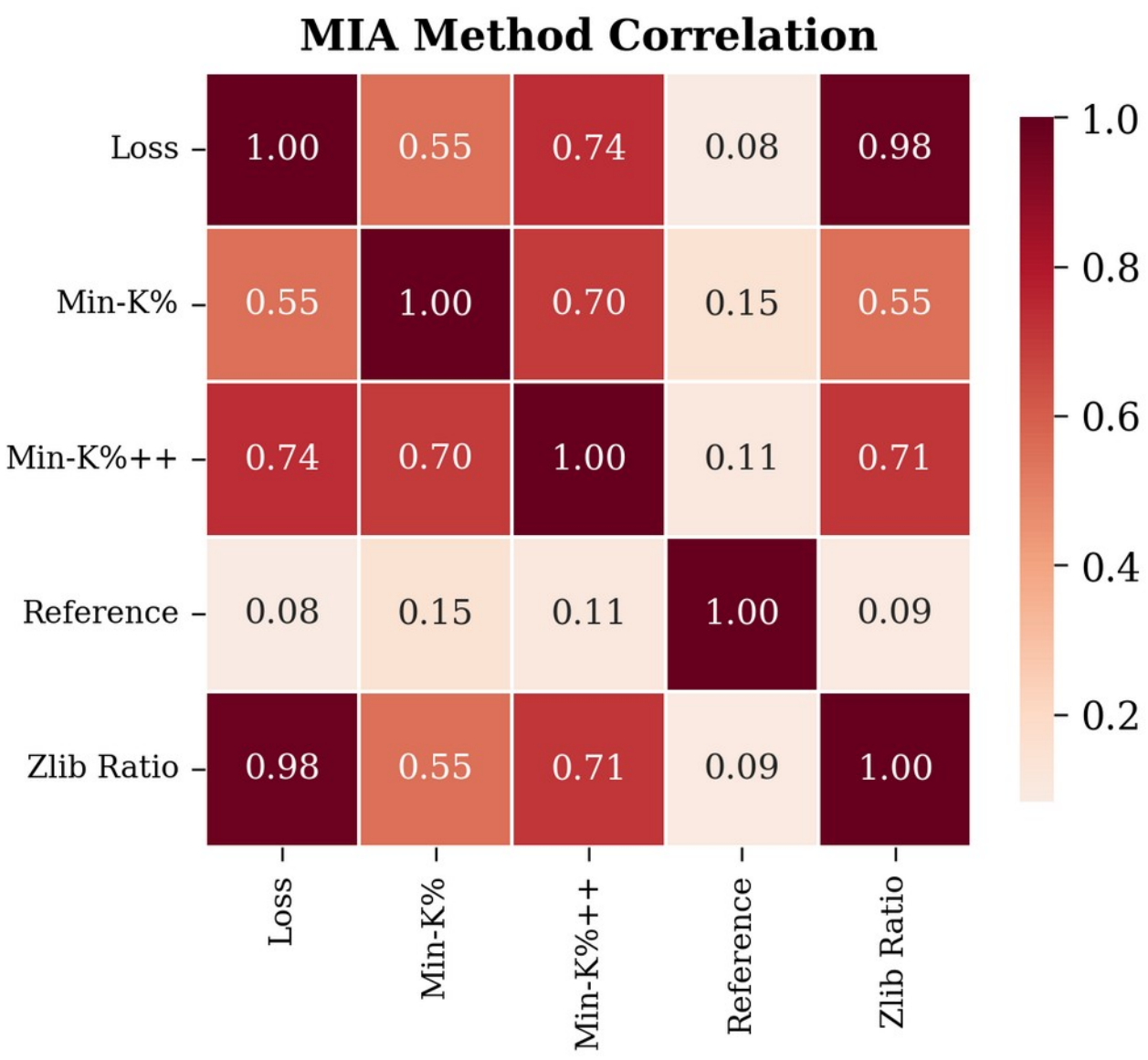

**Fig. 2.** Correlation matrix across five MIA methods. Loss and Zlib are near-redundant (r = 0.98); Reference provides independent information (r < 0.15 with all others).

**MemGuard Composite Score.** Fitted on all seven models, MCS assigns a weight of +62.19 to temporal proximity against −0.42 for mia_loss, +0.22 for mia_zlib, +0.22 for mia_min_k_pp, −0.08 for mia_min_k and −0.03 for mia_ref, all on standardized features. Evaluated in the manner of the original submission, by scoring the same rows the model was fitted to, this yields Cohen's d = 18.57. We do not treat that number as a detection result, and the next two paragraphs set out why, since the reasoning applies to any composite score built the same way.

The first difficulty is that the composite is scored on its own training rows. Table 5 reports five protocols on the same 299,600 prompt-model pairs, ordered from the most permissive to the most conservative. The full composite scored on its fitted rows attains ROC-AUC 1.000; so, however, does temporal proximity on its own, and that is where the second and more fundamental difficulty lies. Because $\tau > 0$ holds exactly when $t < ck$, and in-sample status is defined as $t < ck$, $\tau$ is a monotone transform of the variable that defines the label. Fig. 4 plots the five protocols. Perfect separation is arithmetic. It is not evidence that anything has been detected, and reporting d = 18.57 as a detection result overstates the finding by roughly an order of magnitude. Dropping the temporal feature and cross-validating gives a more defensible picture: the MIA features alone reach AUC 0.895 under random 5-fold and 0.868 when folds are grouped by calendar quarter so that no date appears in both training and test. A restricted variant using only loss and zlib, the two statistics obtainable from a commercial API that exposes completion log-probabilities, reaches 0.865. These are the numbers we regard as the paper's detection results. The pre-cutoff reference corpus offers an independent check on the labelling assumption, and it does not support it uniformly. Comparing Financial PhraseBank against the post-cutoff control prompts, mean loss is lower on the reference corpus for the three GPT-2 models and Phi-2 (Cohen's d = −0.61, −0.63, −0.76 and −0.81), the direction expected if pre-cutoff text is more familiar, but higher for both Qwen models and TinyLlama (+0.79, +0.71 and +0.07). One confound should be stated: the reference corpus is European financial news prose while the controls are templated price prompts, so the two are not distribution-matched and part of the difference is linguistic rather than mnemonic. Even allowing for that, three of seven models run the wrong way, which is why we no longer describe the corpus as guaranteed memorized. Section 5.2 then asks whether even they measure what they appear to.

**Table 5.** Detection quality under five evaluation protocols, on the same 299,600 prompt-model pairs. P0 reproduces the protocol of the submitted version. P1 isolates the temporal feature: it attains a perfect score because $\tau > 0$ holds exactly when $t < ck$, which is the definition of the label. P2 and P3 exclude the temporal feature and cross-validate; P3 groups folds by calendar quarter so that no date appears in both training and test. P4 fits and evaluates within each model, removing between-model scale differences, and reports the range across the seven models.

| Evaluation protocol | ROC-AUC | PR-AUC | TPR at 1% FPR | Folds | Leakage controlled |
|---|---|---|---|---|---|
| **P0 Full MCS, scored on its fitted rows** | **1.000** | **1.000** | **1.000** | **none** | **none** |
| **P1 Temporal proximity alone** | **1.000** | **1.000** | **1.000** | **none** | **none** |
| **P2 MIA features only, random 5-fold** | **0.895** | **0.912** | **0.187** | **5** | **model only** |
| **P3 MIA features only, quarter-grouped** | **0.868** | **0.852** | **0.026** | **5** | **model and date** |
| **P4 Within model, quarter-grouped** | **0.29–0.93** | **0.10–0.99** | **0.00–0.59** | **5** | **model, date, identity** |

What the MIA features contribute once temporal proximity is present has a precise answer. Within a group of models sharing a cutoff, τ is identical for every model, so it cannot order them at all; only the MIA features can. Two such tiers exist in our lineup, the three GPT-2 models and the two Qwen models, and across the 83,100 prompt groups containing more than one model from a single tier the MIA features produce a strict ordering in every case. Whether that ordering is informative is a separate question, and the rest of this section addresses it. A composite score that separates in-sample from out-of-sample prompts is not the same thing as a score that measures memorization, and the two can be told apart. Any evaluation that varies the date confounds memorization with whatever else changes over five and a half years of financial text. So we hold the date fixed and ask the question the partition in Section 3.6 actually depends on: among the seven models scored on the same prompt, can the MIA statistics identify which ones had that date in their training window? Restricting attention to the 33,300 prompt groups in which both classes are present, and fitting nothing, this is a rank comparison within each group. On the raw scores it looks decisive: mean within-group AUC is 0.99 for loss and zlib, 0.86 for Min-K%++, and 0.64–0.67 for the remaining two. The difficulty is that the models differ systematically in the scale of these statistics, a 7B Qwen model assigning lower loss to almost any text than GPT-2 does, and at a given date model identity is close to determining in-sample status. Normalizing within each model removes that shortcut. Under z-scoring, under rank transformation, and under demeaning by model and prompt type, mean AUC across the five methods falls to between 0.487 and 0.537. Three normalizations chosen independently give the same answer, and it is chance. Evaluated per model rather than pooled, five of the seven fall below chance. TinyLlama is the exception, reaching 0.93, and it is worth resolving because it is the only result in this section that looks like successful detection. Detrending each feature on time within the model removes about half its excess over chance, leaving 0.73. A sharper test is available. If memorization drives the separation, it should be clearest near the cutoff, where in-sample and out-of-sample prompts are days apart and differ in little except training exposure; if slow drift drives it, the separation should vanish there, because any smooth trend is nearly constant across a short window. Restricting to symmetric windows around the cutoff, AUC falls from 0.93 on the full sample to 0.79 at twelve months, 0.72 at nine, 0.43 at six and 0.38 at three. It is worst precisely where memorization should be easiest to see, and below chance in the two tightest windows. Phi-2 behaves the same way, falling from 0.65 to 0.15. We therefore attribute the apparent within-model detection to time structure in the MIA statistics rather than to memorization, and we no longer treat TinyLlama as an exception. We read this as showing that what the raw statistics discriminate is which model produced a completion, not whether that completion was memorized.

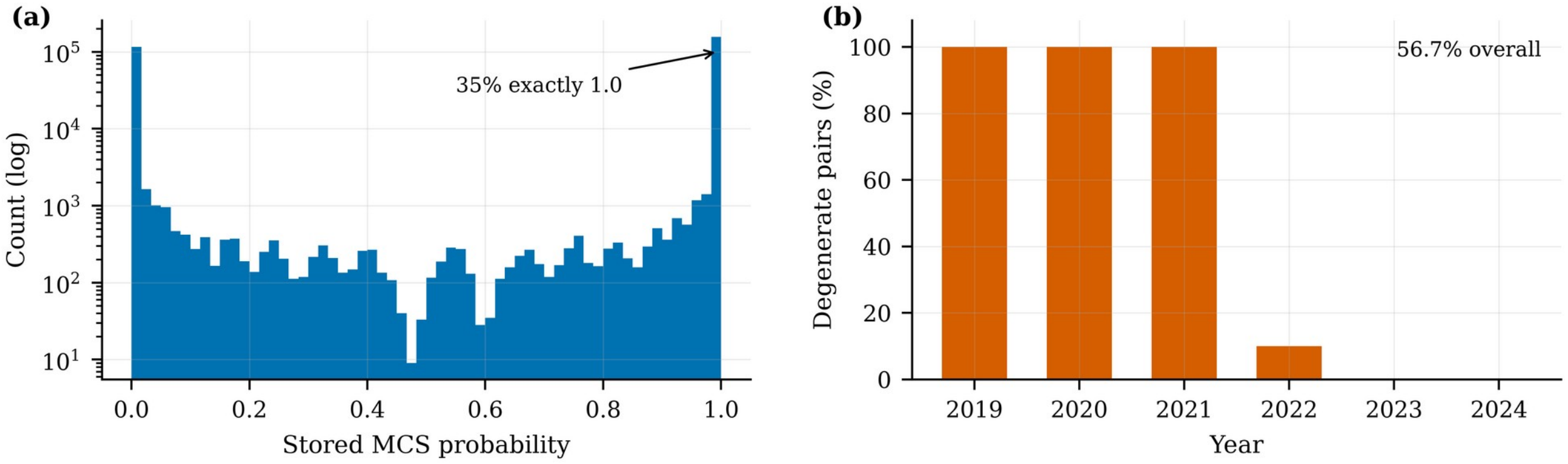

**Fig. 3.** MemGuard Composite Score saturation and partition degeneracy. (a) 35% of stored composite scores equal 1.0 exactly, because the fitted weight on temporal proximity saturates the logistic link in double precision. (b) Share of stock-date pairs on which the per-date median coincides with that saturated value, so every model falls in the clean set and the tainted set is empty. The two conditions correspond exactly across all 13,850 groups.

### *5.2 Portfolio Performance (RQ2)*

**Table 4.** Portfolio performance over 277 sampling dates spanning January 2019 to June 2024, net of 15 bps transaction costs applied symmetrically to every strategy, annualised over the realised 5.48-year span. Excluding the weakest single model outperforms every contamination-based filter. The table also reports short-term reversal, volatility-scaled momentum, sector-neutral and market-neutral momentum, a market-neutral long-short variant of CMMD, and two walk-forward machine-learning forecasters trained on lagged returns, 20-day momentum and realised volatility, refitted every 25 evaluation dates on data strictly preceding each prediction. All five price-based factor strategies lose money over this window. Sector-neutral momentum, formed by demeaning 20-day momentum within GICS sector on each date, attains −1.52; demeaning retains 83% of the cross-sectional dispersion in momentum, so the construction is not degenerate despite three to twelve names per sector, but the sharply lower volatility (2.00% against 6.43%) shows that most of what the unadjusted version earns and loses is a sector bet. Consistent with this, the sector-exposure diagnostic shows CMMD's tilts against equal weight are all below 0.3 percentage points, so CMMD is not a disguised sector position. Gradient boosting attains a hit rate of 52.8% and a Sharpe of 0.22, indistinguishable from the unfiltered LLM ensemble and below the trivial model-exclusion baseline, which is itself a useful calibration: whatever the LLM signals contribute here is not more than a conventional classifier extracts from price history alone.

| Strategy | Total Return | Ann. Return | Ann. Volatility | Sharpe | Max Drawdown |
|---|---|---|---|---|---|
| **EW Buy-Hold** | **55.63%** | **8.40%** | **8.73%** | **0.96** | **−13.59%** |
| **Raw Alpha minus Qwen-7B** | **10.14%** | **1.78%** | **4.63%** | **0.38** | **−9.59%** |
| **CMMD, rank split (corrected)** | **10.50%** | **1.84%** | **4.99%** | **0.37** | **−10.21%** |
| **Debiased Alpha** | **3.97%** | **0.71%** | **2.40%** | **0.30** | **−5.15%** |
| **Raw Alpha (7 models)** | **4.33%** | **0.78%** | **3.54%** | **0.22** | **−8.47%** |
| **ML gradient boosting** | **8.27%** | **1.46%** | **6.70%** | **0.22** | **−11.82%** |
| **CMMD, published median split** | **4.40%** | **0.79%** | **4.16%** | **0.19** | **−10.17%** |
| **GPT-2 trio only** | **3.66%** | **0.66%** | **4.45%** | **0.15** | **−10.57%** |
| **Strict-OOS ensemble** | **3.47%** | **0.63%** | **4.62%** | **0.14** | **−10.57%** |
| **ML logistic regression** | **0.07%** | **0.01%** | **7.54%** | **0.00** | **−14.06%** |
| **Momentum-20d** | **−15.20%** | **−3.00%** | **6.43%** | **−0.47** | **−16.63%** |
| **Vol-scaled momentum** | **−21.57%** | **−4.33%** | **5.88%** | **−0.74** | **−19.02%** |
| **Reversal-5d** | **−24.59%** | **−5.02%** | **6.22%** | **−0.81** | **−20.14%** |
| **Sector-neutral momentum** | **−15.36%** | **−3.04%** | **2.00%** | **−1.52** | **−16.02%** |
| **Market-neutral momentum** | **−26.07%** | **−5.44%** | **2.53%** | **−2.15** | **−26.07%** |
| **CMMD long-short (market-neutral)** | **−14.45%** | **−2.81%** | **0.59%** | **−4.78** | **−14.45%** |
| **Random** | **−33.37%** | **−7.14%** | **1.59%** | **−4.50** | **−34.08%** |

Two implementation defects in our original backtest have to be reported before these numbers can be read. First, transaction costs were applied to Raw Alpha and Debiased Alpha but not to CMMD, which was scored gross. CMMD's turnover is 0.38 against Raw Alpha's 0.31, so it should in fact pay slightly more. Second, the 93 sampling dates of the original evaluation are roughly 21 calendar days apart across a 5.48-year span, and annualising them as though they were consecutive trading days inflated every Sharpe ratio in the table by a factor of about 4.5. Table 4 above is computed with costs applied symmetrically, annualised over the realised span, and extended to all 277 dates for which MIA scores exist. On that basis CMMD attains a Sharpe of 0.37 against 0.22 for the unfiltered ensemble, and simply excluding Qwen-7B, the one model with negative mean returns in both regimes, attains 0.38 on lower turnover.

The finding that threshold debiasing underperforms the unfiltered ensemble survives correction and is reported in Section 5.7 across the full percentile sweep. What does not survive is the claim that CMMD improves on the unfiltered ensemble. A two-sided block bootstrap on the Sharpe difference gives $p = 0.016$, but a paired t-test on the same daily differences gives $p = 0.114$; when two tests of the same comparison disagree this sharply at $n = 277$, the honest reading is that the effect is not robust to the choice of statistic. Excluding Qwen-7B, by contrast, is significant on both (bootstrap $p < 0.001$, paired $p = 0.045$) and produces a higher Sharpe than any contamination-based filter we test.

The submitted version reported a paired bootstrap on this comparison giving a mean Sharpe difference of +1.35 with a 95% interval of [−0.28, +3.12] and $p = 0.054$, followed by a one-sided $p = 0.027$. We withdraw that result. It does not reproduce from our released artifacts, the +1.35 is the arithmetic difference of two point estimates rather than a bootstrap mean, and switching to a one-sided test after a two-sided result narrowly missed significance is not defensible when the direction was not pre-specified. All tests reported here are two-sided, computed on a stationary block bootstrap with block length five, and reproducible from the released code.

The economic reading changes accordingly. Regressing each strategy's daily returns on the Fama-French five factors plus momentum (Fig. 12), with HC3 robust standard errors and alpha reported in basis points per day rather than annualised, no LLM-derived strategy attains significant factor-adjusted alpha: CMMD gives +3.32 bps/day at $t = 0.72$, the unfiltered ensemble +0.86 at $t = 0.28$, and excluding Qwen-7B +3.05 at $t = 0.73$. We use HC3 rather than a heteroskedasticity- and autocorrelation-consistent estimator because the observations are irregularly spaced and the measured first-order autocorrelation is approximately zero, so a lag structure would be arbitrary. The market-neutral variant, which removes the net long exposure, produces −6.56 bps/day at $t = -12.85$. Whatever the unfiltered signals earn over this period is compensation for market exposure, and filtering does not change that.

The gap between the equal-weight benchmark and every LLM strategy still requires comment. CMMD is long-only and signal-weighted, so it carries substantial net long exposure and its returns load on market beta; the correct like-for-like comparison is therefore against the unfiltered ensemble, which shares that exposure, rather than against buy-and-hold. On that comparison CMMD attains 0.37 against 0.22, a difference that does not survive both of the tests we apply to it. We also report the market-neutral variant that the submitted version deferred as out of scope: cross-sectionally demeaning the clean signal each date removes the beta exposure and produces −5.63 bps/day, with a factor-adjusted alpha of −6.56 bps/day at $t = -12.85$. Once market exposure is stripped out, what remains is significantly loss-making.

The comparison between threshold debiasing and the unfiltered ensemble survives correction, though at a smaller magnitude: 0.30 against 0.22 on the corrected series, with the bootstrap interval spanning zero ($p = 0.38$). The interpretation we offered originally still holds and is worth retaining. Detecting contamination and acting on it in a portfolio are different problems. A contamination score can order signals by how memorization-driven they are and still be useless as a filter, because a memorized signal may be directionally correct and zeroing it discards a profitable position along with the bias.

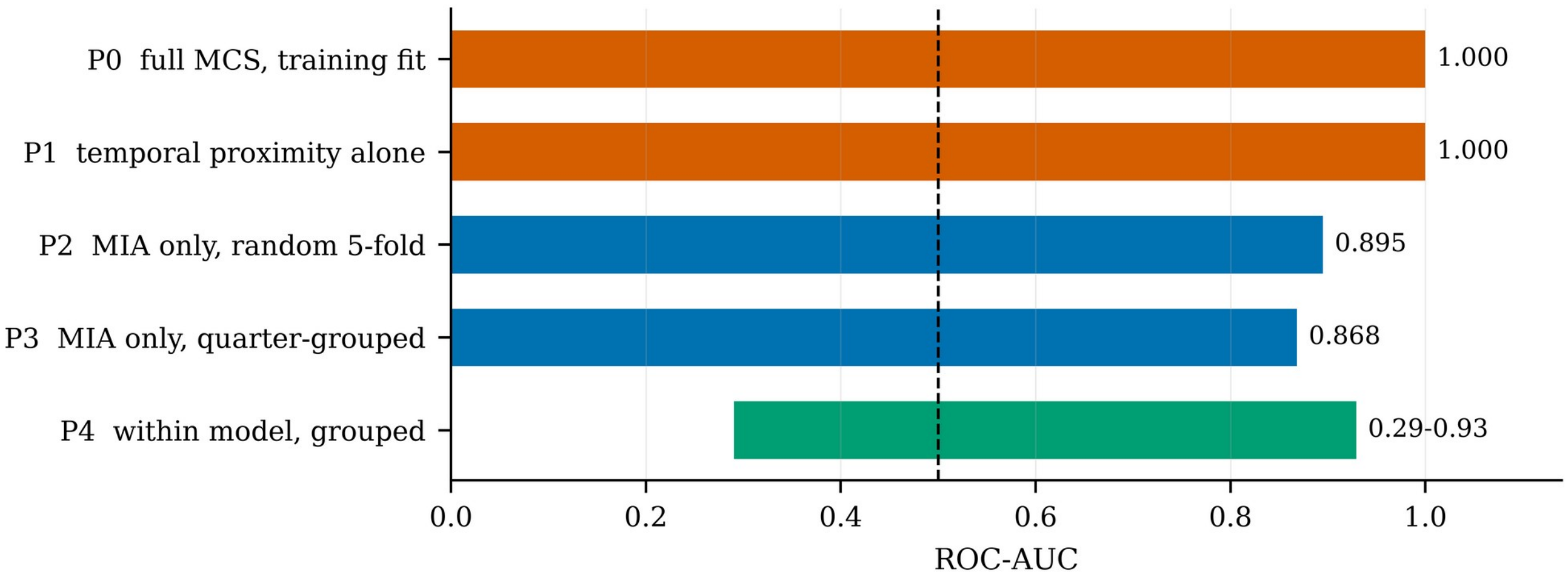

**Fig. 4.** Detection quality under five evaluation protocols (Table 5). Perfect separation under P0 and P1 is arithmetic rather than detection, since the temporal feature is a monotone transform of the variable that defines the label. Dashed line marks chance.

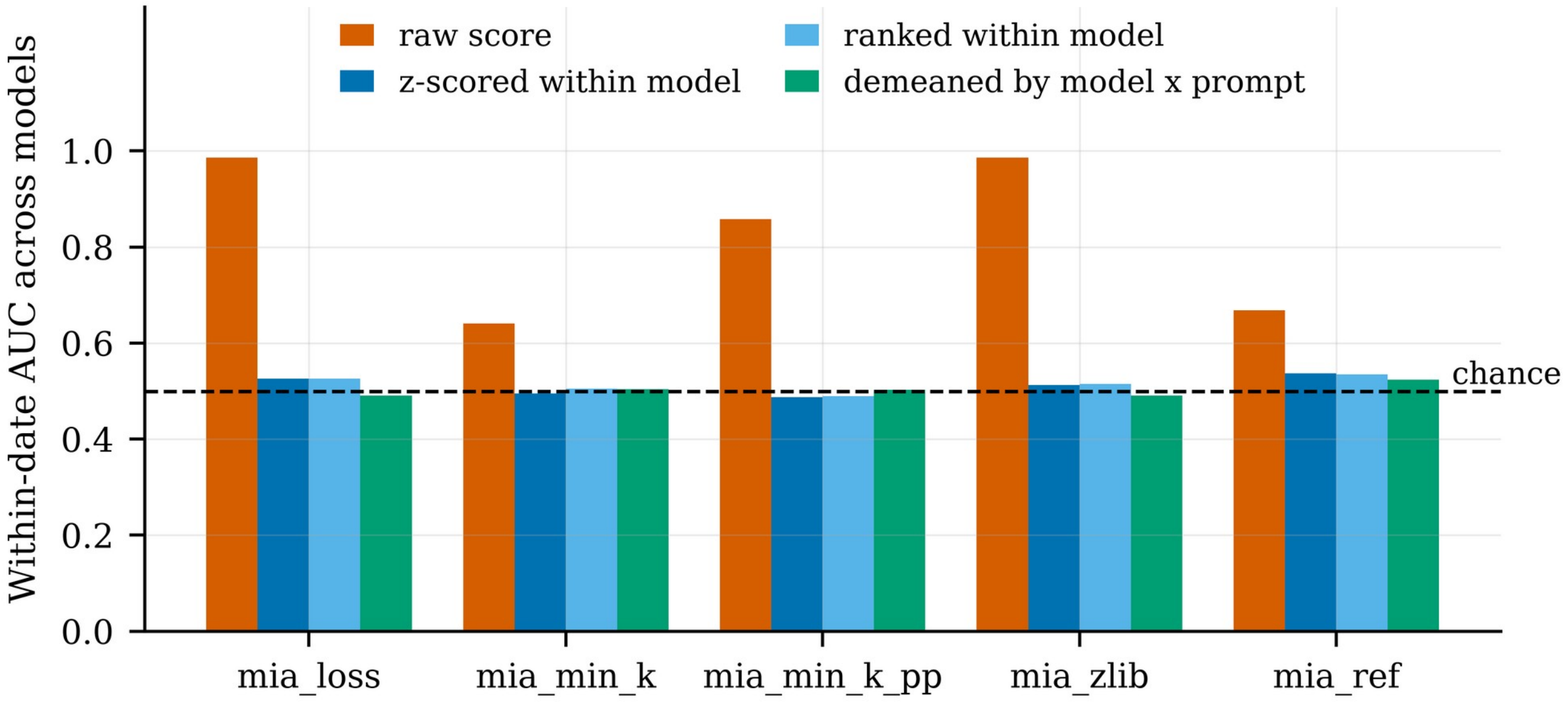

**Fig. 5.** Within-date, across-model discrimination. At a fixed date, can MIA scores identify which models had that date in training? Date is held constant and no model is fitted, so neither temporal leakage nor overfitting is possible. Raw scores appear informative, but models differ systematically in baseline score scale. Under three independent within-model normalizations discrimination falls to 0.487–0.537, computed over 33,300 label-varying groups. Dashed line marks chance.

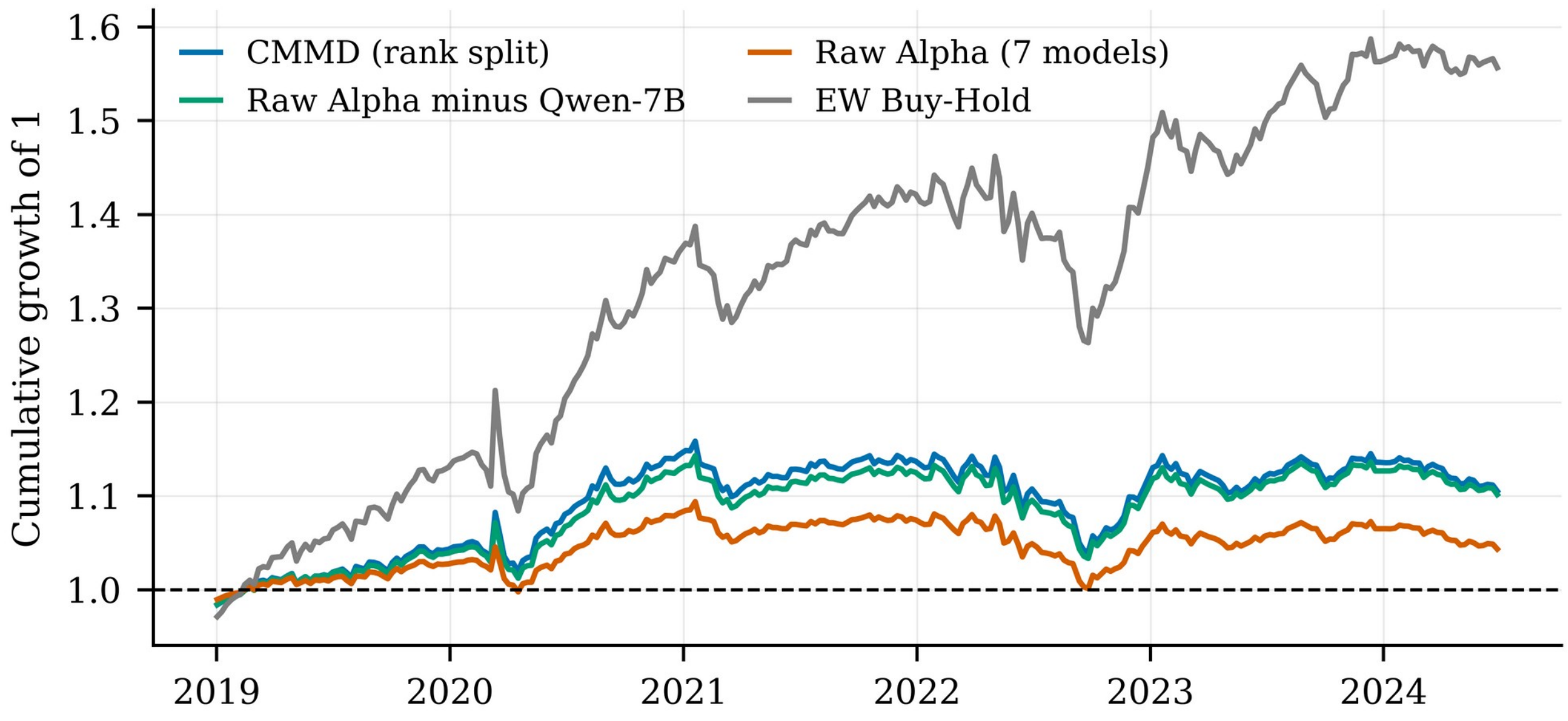

**Fig. 6.** Cumulative returns over 277 sampling dates, net of 15 bps transaction costs applied symmetrically to every strategy. The equal-weight basket dominates every LLM-derived strategy throughout.

### *5.3 Signal Accuracy vs. Contamination (RQ1/RQ2)*

Fig. 7 relates directional accuracy to contamination, measured as quintiles of mia_loss. This is the quantity the evaluation code actually used; the composite score gives a similar picture and both are reported. In-sample accuracy rises across the range, from 39.7% in Q1 to 45.9% in Q5. Out-of-sample accuracy does not fall monotonically, which the submitted version reported that it did: it is 34.6% in Q1, rises to 41.7% in Q2, and then drifts down to 41.1% in Q5. Q1 is the lowest point of the entire series, in-sample or out, and quoting the out-of-sample range from Q2 onwards, as we previously did, conceals that. The two curves cross between Q2 and Q3, so a crossover exists, but it is weaker than reported and it is not the clean reversal the original description implied.

There is a further difficulty with reading the pooled curves as evidence about memorization at all. The in-sample pool and the out-of-sample pool are not drawn from the same models. Because coverage is determined by each model's cutoff, the in-sample pool is 23.5% Qwen-3B and 21.6% Phi-2 but only 3.4% GPT-2, while the out-of-sample pool is 28.8% GPT-2 and 2.0% Qwen-3B. Since the models differ in baseline directional accuracy by up to thirty percentage points, a substantial part of the vertical gap between the two pooled curves reflects which models populate each pool rather than any effect of contamination. Panel (b) of Fig. 7 recomputes the relationship within each model, where that confound cannot operate. The within-model gaps run from −0.6 to +4.7 percentage points, an order of magnitude smaller than the pooled comparison suggests, and they do not rise systematically with contamination.

The practical implication is narrower than the submitted version claimed. A researcher who evaluated LLM-generated forecasts only on dates inside the training window would see accuracy rising with familiarity and could mistake that for skill. That much holds. But the out-of-sample curve does not fall monotonically, the crossover is confined to the middle quintiles, and the within-model gaps are of the order of a few percentage points rather than the reversal the pooled curves suggest. We therefore present the crossover as a reason for caution in evaluation design, not as a quantitative diagnostic, and we withdraw the recommendation that practitioners monitor it as an operational signal.

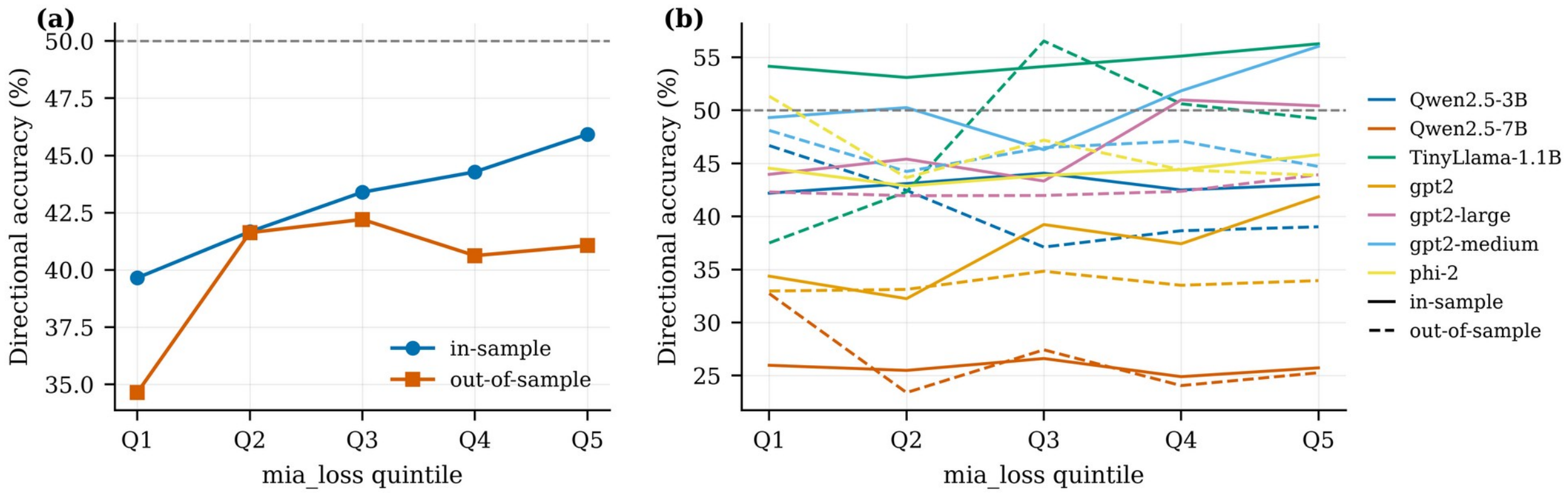

**Fig. 7.** Directional accuracy against contamination quintile. (a) Pooled across models: the out-of-sample curve is not monotone, and Q1 is the lowest point of the series. (b) The same relationship computed within each model. Because coverage is set by each model's cutoff, the in-sample and out-of-sample pools are drawn from different models, so the pooled comparison confounds contamination with model identity. Neutral signals count as incorrect, following the evaluation code. Dashed line marks a 50% hit rate.

## *5.4 Scale and Temporal Effects (RQ3)*

Figs. 8 and 9 disaggregate by model and by prompt type. Within the GPT-2 family (same cutoff, different sizes), separation increases with parameters: $d = -0.19$ (124M), −0.26 (355M), −0.33 (774M), confirming that larger models memorize more [20]. Across families, TinyLlama (1.1B, Sep 2023) shows stronger separation ($d = -1.37$) than GPT-2 Large (774M, Oct 2019) despite being only 1.4x larger, demonstrating that training data recency dominates model size. Qwen 3B and 7B (same cutoff) show nearly identical $d = -0.39$ and −0.37, suggesting a saturation effect.

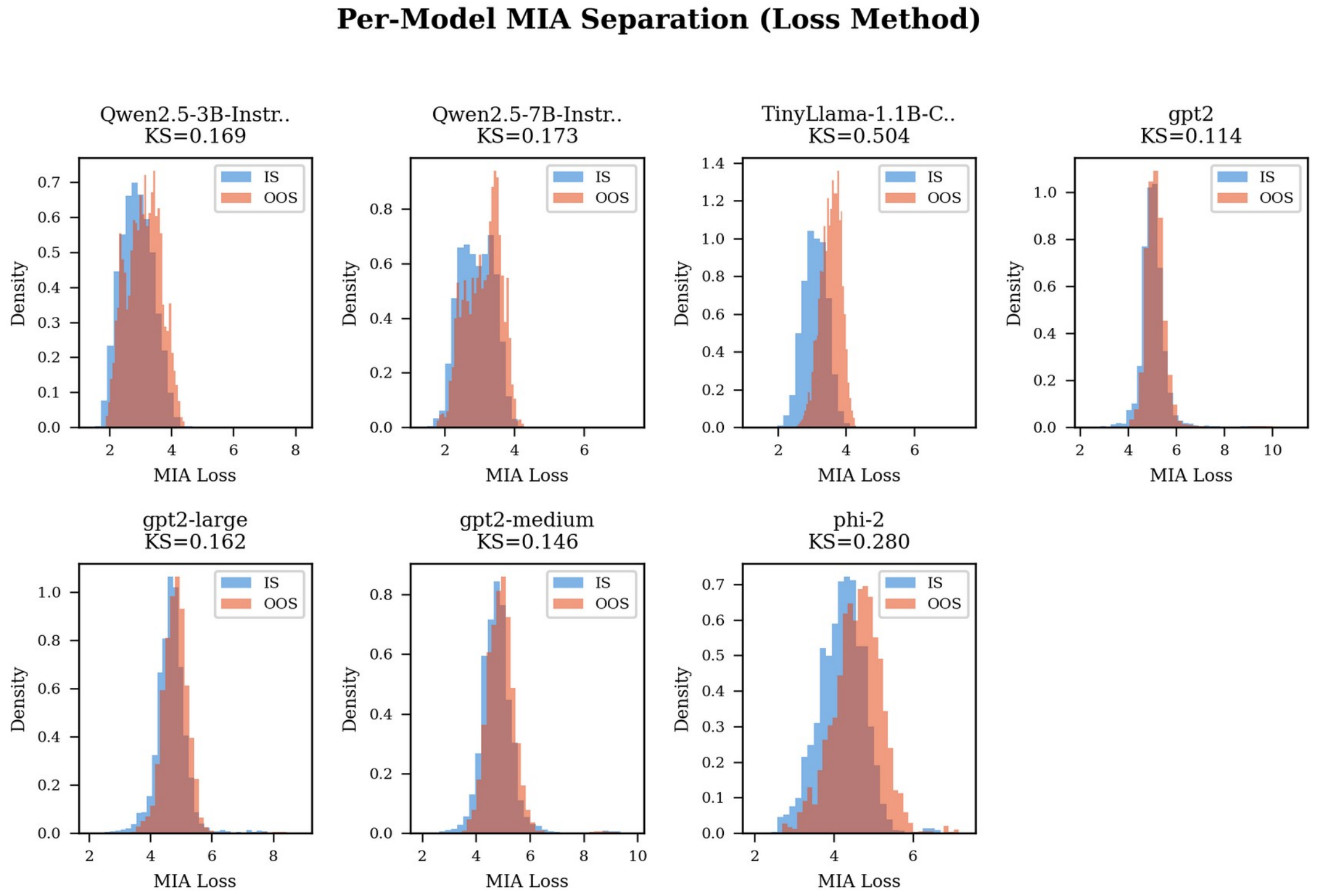

**Fig. 8.** Per-model MIA separation using the loss method. Models with recent training cutoffs (TinyLlama, Phi-2) show stronger IS/OOS divergence than older models (GPT-2 family).

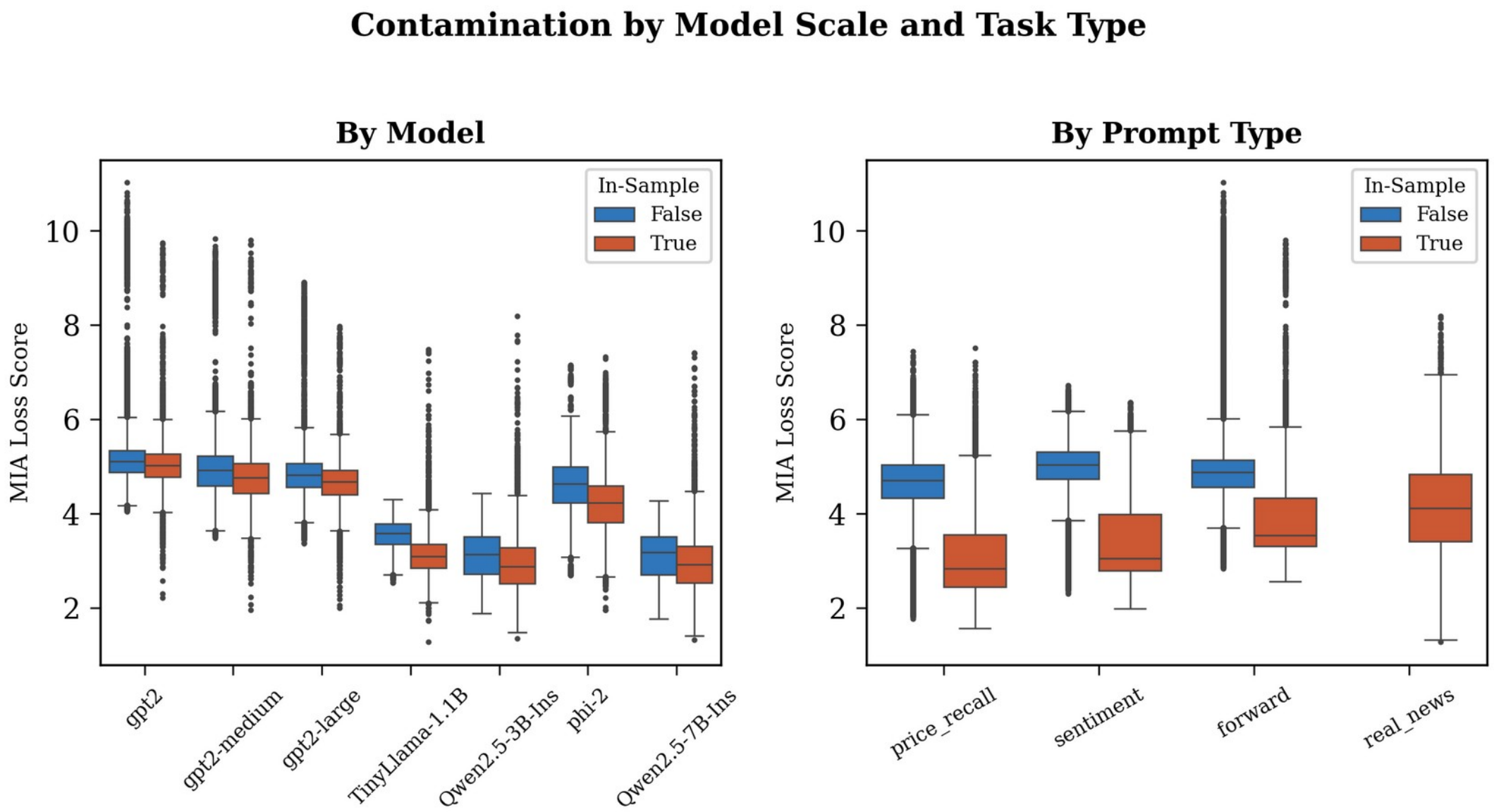

**Fig. 9.** MIA contamination scores by model (left) and prompt type (right), split by IS/OOS status. Sentiment prompts show the strongest separation (d = −2.05 on mia_loss), marginally ahead of price recall (−1.99); forward-looking prompts separate least (−1.47).

## *5.5 Per-Model Signal Analysis*

Disaggregating by model gives a much weaker picture than the submitted version reported, and in one respect the opposite one. Under the accuracy rule the evaluation code applies, in which neutral signals count as incorrect, the in-sample minus out-of-sample gaps span −0.6 to +4.7 percentage points across the seven models, and the GPT-2 family runs in the same direction as the rest rather than inverting. Both differ from the submitted version, which reported gaps of twenty to thirty-five points and an inverted GPT-2 pattern, and offered a market-volatility explanation for an inversion that is not present in the data. With between 850 and 1,900 out-of-sample observations per model, none of these gaps is individually distinguishable from zero, so we report the range and draw no per-model inference from it.

The GPT-2 family runs in the same direction as the others, not the inverse, which is the reverse of what the submitted version stated: in-sample accuracy is 36.7%, 50.1% and 46.2% for the base, medium and large models against out-of-sample values of 33.7%, 46.0% and 42.5%, so gaps of 3.0, 4.1 and 3.6 points in favour of in-sample. The explanation previously offered for the supposed inversion, that early-2019 in-sample dates coincided with post-selloff volatility, was constructed to account for a pattern that is not present in the data. Note also that Qwen and Phi-2 have only 850 and 1,900 out-of-sample observations respectively, so their gaps carry standard errors of several percentage points and none of the four is individually distinguishable from zero.

Qwen-7B is worth separating out. It is the largest model in the lineup yet produces a markedly bearish signal distribution (13.1% bullish, 40.6% bearish, 46.3% neutral) and is the only model whose signals lose money on average in-sample (−8.17 bps/day across 13,000 observations); out-of-sample it is close to flat (+0.96 bps across 850). Because the contamination ordering places it in the tainted group on essentially every non-degenerate date, CMMD excludes it by default, and Section 5.8 shows that excluding it explicitly outperforms the filter. Its behaviour most likely reflects instruction tuning toward caution rather than any property of memorization, which is itself a caution against reading the clean-tainted contrast as a memorization effect.

Signal distributions differ sharply across models. TinyLlama is almost always bullish (87.3% bullish, 1.8% neutral); GPT-2 base is the most balanced (48.2% bullish, 35.7% neutral); Qwen-7B is the outlier described above. This dispersion is what allows a partition to produce different clean and tainted signals at all. Under the corrected rank split, 2,854 of the 13,850 stock-date pairs show clean and tainted groups pointing in opposite directions, and 5,602 (40.4%) show a disagreement magnitude above 0.5. Under the published median split the corresponding counts are 1,213 and 2,444, computed over the 6,000 pairs where a non-degenerate partition exists; the remaining 7,850 pairs have an empty tainted set and are excluded, since on those the disagreement is by construction the negative of the clean signal. The submitted version reported 1,863 of 4,650 pairs (40.1%) without that exclusion. This resolves the inconsistency between the two counts noted in review: the smaller figure counts sign conflicts, the larger counts disagreement magnitude above a threshold, and they are not alternative estimates of the same quantity.

Returns are unevenly distributed across the cross-section. On the 277-date series the strongest contributors under the corrected rank split are TSLA (+33.74 bps/day), HON (+19.81), LLY (+19.20), XOM (+18.84) and BAC (+18.27); the weakest are SBUX (−4.49), NKE (−3.12), JNJ (−2.16) and MRK (−0.23). We previously read the concentration in high-volatility, heavily covered names as consistent with a memorization mechanism. We no longer draw that inference. The composition changes substantially between the 93-date and 277-date samples, which is what one expects of idiosyncratic variation in a 50-stock cross-section rather than of a stable mechanism, and with 277 observations per stock these per-name estimates carry standard errors of the same order as the estimates themselves.

## 5.6 Strategy Correlation Analysis

Return correlations among the strategies are informative about what filtering actually changes. On the 277-date series CMMD correlates at 0.984 with the unfiltered ensemble, 0.996 with the ensemble excluding Qwen-7B, and 0.990 with the equal-weight basket; the unfiltered ensemble correlates at 0.969 with equal-weight. Every LLM-derived strategy is therefore close to a market-beta position with minor perturbations, which is consistent with the factor regressions in Section 5.2, where the market loading is the only reliably estimated coefficient. Filtering changes which perturbation is applied; it does not produce a strategy with a materially different return profile.

The correlation of 0.984 between CMMD and the unfiltered ensemble is itself informative. Under the published median split the two are not merely similar but identical on 56.7% of stock-date pairs, because the partition degenerates there and every model falls in the clean set. On the remaining pairs the clean-model consensus earns 9.58 bps/day against 2.01 for the tainted group, a ratio of 4.8 (paired $t = 6.57$, $p < 0.001$) under the corrected rank split. The submitted version reported this contrast as sevenfold, 14.48 against 2.13 bps, but that figure was computed across all pairs including the degenerate ones, where the tainted signal is identically zero by construction and contributes nothing to the tainted mean. The contrast is real; it is smaller than we reported, and it does not translate into portfolio performance.

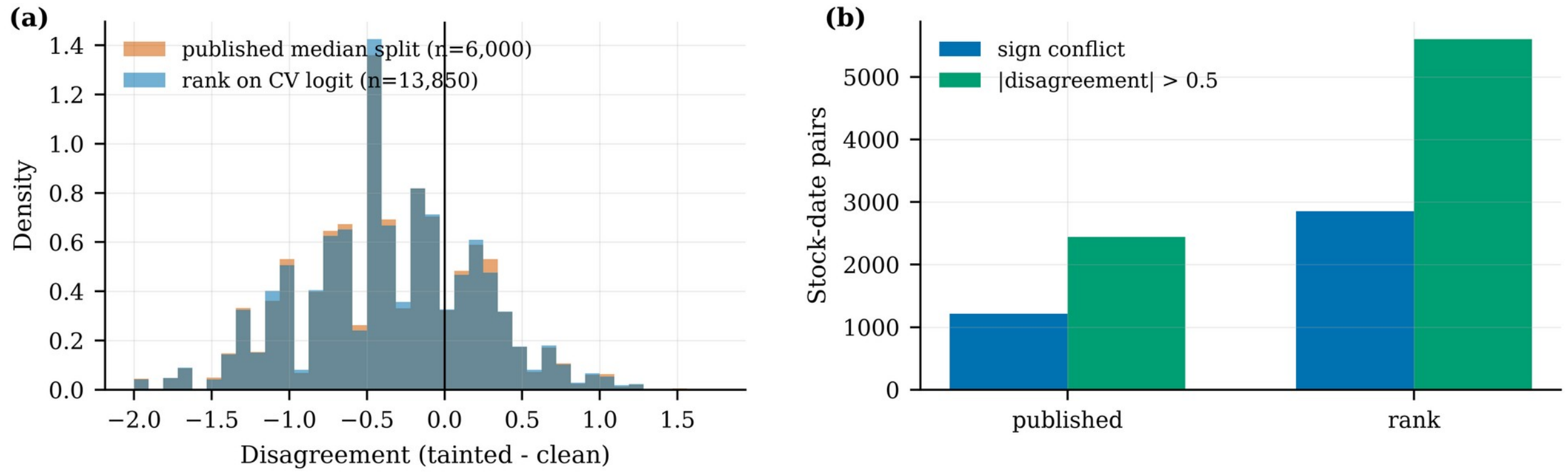

**Fig. 10.** Cross-model disagreement. (a) Distribution of the tainted-minus-clean signal difference under the published median split and under the corrected rank split. (b) The two disagreement metrics reported in the submitted version, computed on stock-date pairs where a non-degenerate partition exists.

## 5.7 Threshold Robustness (RQ4)

Fig. 11 repeats the threshold sweep with costs applied symmetrically and on both contamination scores: the raw loss statistic used by the evaluation code and the composite score described in Section 3.7. The conclusion of the submitted version survives and is if anything strengthened. At aggressive thresholds (P10–P25) most signals are zeroed and what remains carries little information; at permissive thresholds (P75–P95) almost nothing is filtered and the strategy converges on the unfiltered ensemble. No threshold on either score improves on the unfiltered baseline, and filtering on the composite score is uniformly worse than filtering on the raw loss. This was the finding that originally motivated a threshold-free design; it remains valid, but Section 5.2 shows that the threshold-free alternative does not solve the problem either.

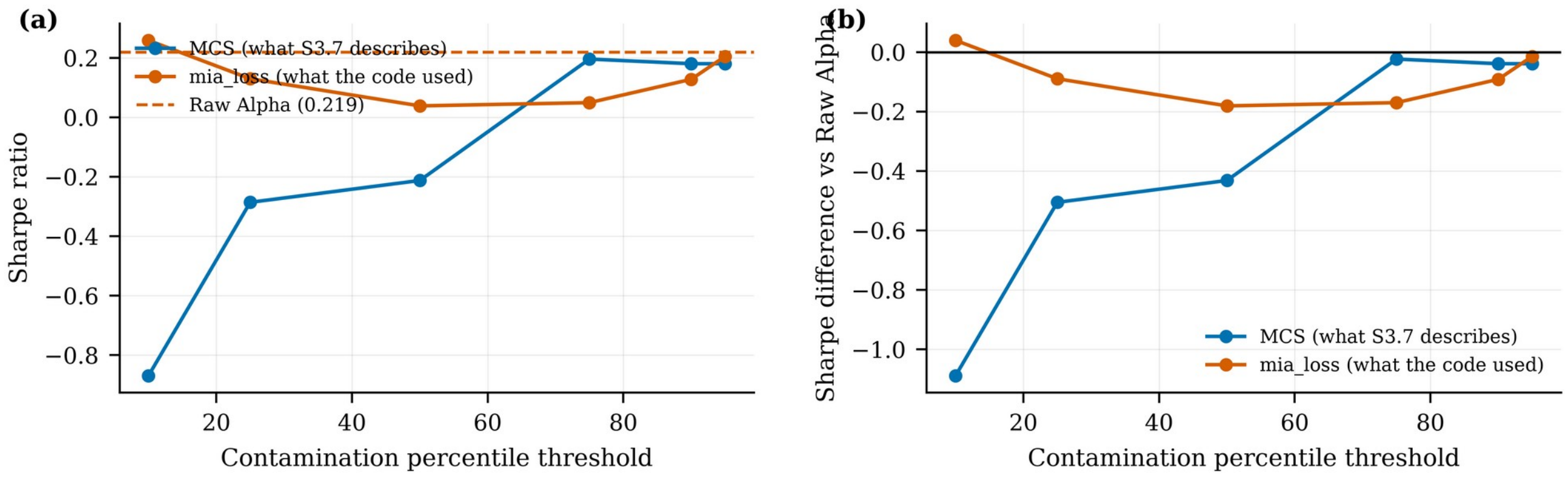

**Fig. 11.** Threshold sensitivity of contamination filtering, on the raw loss score used by the evaluation code and on the composite score described in Section 3.7, with costs applied symmetrically. (a) Sharpe ratio against threshold percentile, with the unfiltered ensemble marked. (b) Difference against the unfiltered ensemble. No threshold on either score improves on the baseline.

### *5.8 Ablation Study: MCS Component Contributions*

We ablate the composite score over thirteen feature subsets, in every case under quarter-grouped five-fold cross-validation rather than a training fit. Any variant retaining temporal proximity attains AUC of 1.000 or 0.999, for the reason given in Section 5.1. Among the variants that exclude it, all five MIA features together reach 0.868, loss alone 0.859, zlib alone 0.839, the reference method 0.802, Min-K%++ 0.738 and Min-K% 0.578. Dropping any single MIA feature from the full set changes almost nothing, which is consistent with the high correlation between loss and zlib reported in Figure 2.

We had drawn a distinction here between detection, which we said relies on temporal proximity, and debiasing, where we argued the MIA features add value because they differentiate models at a fixed date while τ cannot. The first half stands. The second does not: Section 5.1 shows that the within-date ordering the MIA features produce is at chance once between-model scale differences are removed, and Section 5.2 shows the portfolio benefit that ordering was supposed to deliver does not appear.

A second ablation varies the partition rule itself, and it is the more informative of the two. Splitting the seven models at random into halves, repeated with ten seeds, gives a Sharpe of −0.02 with a standard deviation of 0.05, well below the unfiltered ensemble at 0.22. The partition is therefore not arbitrary: a random split destroys the signal. But the corrected rank split reaches only 0.37, the published median split 0.19 and a soft contamination-weighted average 0.17, none of which exceeds the 0.38 obtained by excluding Qwen-7B. Leaving each model out in turn moves CMMD between 0.15 and 0.41, with the highest value obtained by excluding GPT-2 base. Taken together, the ordering of models by contamination score carries information, but not information that survives being turned into a portfolio.

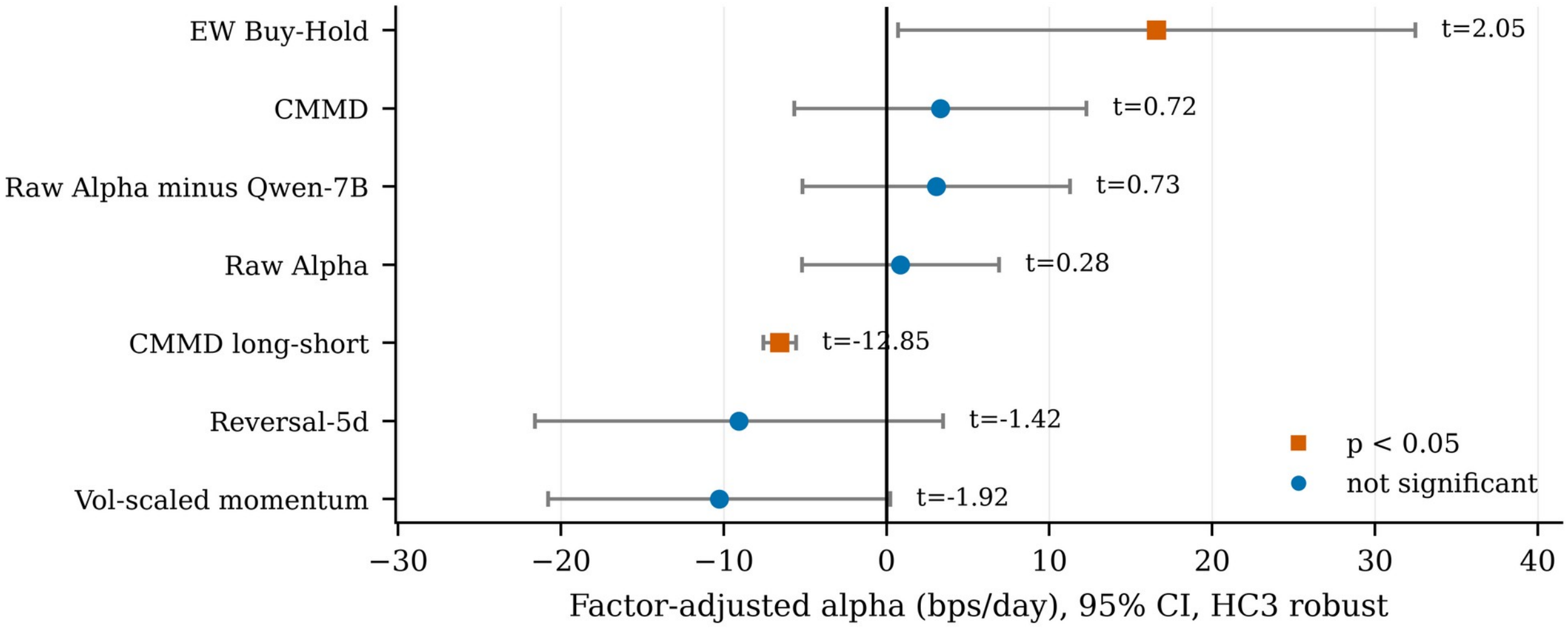

**Fig. 12.** Factor-adjusted alpha. Intercepts from regressions on the Fama-French five factors plus momentum, with HC3 robust standard errors and 95% confidence intervals, reported in basis points per day and not annualised. HC3 rather than a HAC estimator because observations are irregularly spaced and measured first-order autocorrelation is approximately zero. No LLM-derived strategy attains significant factor-adjusted alpha. Cumulative returns net of symmetric costs are shown in Fig. 6, the disagreement decomposition in Fig. 10, the threshold sweep in Fig. 11, and the minimum-detectable-effect curve in Fig. 13.

# 6. Discussion

## *6.1 The Case for Detection Over Prevention*

Our results contribute to a growing and consequential debate about how to handle memorization in financial LLMs. Three competing paradigms have emerged, each with distinct trade-offs.

**Prevention through temporal retraining** (ChronoBERT [9], PiT-Inference [6]) produces clean models but at prohibitive cost. Training a single frontier-quality LLM requires millions of dollars and months of compute. Moreover, the proliferation of LLM releases means that any temporally retrained model quickly becomes outdated as newer, more capable base models are released. Practitioners face a choice between temporal cleanliness and model quality. Post-hoc filtering appears to sidestep that trade-off by operating on any available model, which is what motivated this work; our results indicate it does not, because the score on which the filtering depends does not measure what it needs to. Read alongside He et al. [9], the two literatures converge from opposite directions. They build chronologically clean models and find the performance gap against a contaminated model to be modest, which bounds the size of the prize. We attempt to detect and remove contamination post hoc and find the detection signal is not reliable enough to act on. A small prize and an unreliable instrument are a poor combination, and together they suggest that effort in this area is better directed at evaluation design than at filtering.

**Prevention through anonymization** (entity-neutering [10], BlindTrade [11]) is computationally cheap but empirically costly. Wu et al. [12] demonstrated that anonymization can destroy more signal than it removes bias, particularly when numerical data (prices, volumes, ratios) and entity relationships (sector peers, supply chain connections) are stripped from the input. Filtering at the signal level does preserve all information content, since the full financial context reaches the model unaltered; this remains an advantage of the approach in principle. What our results show is that preserving the information is not sufficient if the score used to filter cannot distinguish memorized signals from the rest.

**Detection and filtering** treats memorization as an unavoidable property of models trained on internet-scale corpora and attempts to manage its effect at the level of signals. The motivating observation remains sound: memorization is not uniformly harmful, and a model that has learned how rates affect bank equities may reason well despite having memorised those prices. The premise of the approach is that a contamination score can separate the two cases at the level of an individual signal. Our results indicate it cannot, at least with the membership-inference statistics available for open-weight models, which is the central negative finding of this paper.

### *6.2 Why Simple Debiasing Fails*

On the corrected series threshold filtering attains 0.30 against 0.22 for the unfiltered ensemble, a difference well within sampling error, and no threshold across the P10–P95 sweep improves on the baseline (Section 5.7). Three mechanisms account for why a filter that correctly identifies contamination can still fail to help.

**First, memorization is heterogeneous across stocks.** The first method shows how heterogeneity exists within each stock's memorization. While the model has memorized a great deal of AAPL related content, the signal for AAPL will likely contain much contamination from this information. A thinly covered name such as ACN may carry a high contamination score for reasons that have nothing to do with memorised price history, for instance because its prompts are linguistically unusual rather than familiar. Allowing the same filter to remove the signal for AAPL does not treat these two cases similarly.

**Second, some memorized signals are directionally correct by coincidence.** The second method demonstrates how some of the memorized signals are directionally correct merely by chance. For example, a model may have memorized that NVDA rose in 2021. If the model is asked to forecast NVDA's direction on a given date in 2021 and produces a "bullish" forecast, then even though this forecast is contaminated, it is still directionally correct. Thus, if one were to eliminate this forecast (as is done when applying a filter), one would also be eliminating a potentially profitable trade. A consensus rule of the CMMD kind avoids discarding such a signal outright, since it averages rather than zeroes; whether the resulting average is better is the question Section 5.2 answers, and the answer is that it is not reliably so.

**Third, aggressive filtering reduces effective diversification.** Finally, the third method describes how extreme filtering can cause the portfolio to become concentrated in fewer positions. As such, the portfolio is exposed to greater levels of idiosyncratic noise at the level of individual stocks. A consensus rule avoids this by construction, since it produces a signal for every stock-date pair rather than zeroing any. This is a genuine advantage over threshold filtering and it does not depend on the contamination score being informative.

### *6.3 Interpreting MCS Feature Weights*

The weight on temporal proximity (+62.19, against +0.22 to −0.42 for every MIA feature) is not a finding about memorization but a consequence of how the label is defined, as Section 5.1 sets out. Its practical implication concerns deployment rather than interpretation. Where cutoff dates are published, as most commercial providers now do, the temporal feature is redundant with information the practitioner already has and adds nothing a date comparison would not. Where they are unknown, the MIA features are all that remain, and a variant restricted to loss and zlib, the two statistics obtainable from an API exposing completion log-probabilities, reaches AUC 0.865 under grouped cross-validation. We report that as the realistic ceiling for API-based deployment, superseding the estimate of d = 0.2–0.5 given in the submitted version, which was not computed.

### *6.4 Practical Deployment Guidelines*

We are not in a position to recommend deploying contamination-based filtering, and the guidance we offered in the submitted version is withdrawn. What our results do support is a set of cautions for anyone evaluating such a scheme. First, do not evaluate a contamination score against labels defined by the same cutoff dates the score consumes; the separation this produces is arithmetic (Section 5.1). Second, normalize within model before comparing scores across models, since between-model differences in scale dominate the raw statistics (Section 5.2). Third, check the numerical range of any bounded score before splitting it at a sample statistic, since saturation silently degenerates the partition (Fig. 3). Fourth, apply transaction costs to every arm of the comparison and annualize over the realised sampling calendar, not the nominal trading year. Fifth, benchmark against the trivial alternative of excluding the weakest model in the ensemble, which outperformed every filter we tested. Sixth, compute the minimum detectable effect before interpreting a portfolio-level result; at the scale of this study it is roughly 3.9 bps/day.

### *6.5 Computational Cost and Scalability*

A practical deployment consideration is the computational overhead of the MemGuard-Alpha pipeline. The pipeline consists of three stages with distinct cost profiles. Stage 1 (MIA scoring) requires one forward pass per model per prompt for loss-based metrics, plus one additional forward pass using the reference model. In our experiments, scoring 299,600 prompt-model pairs across seven models required approximately 4.5 hours on a single RTX 5090 GPU, translating to roughly 54 milliseconds per prompt-model pair. For comparison, alpha signal generation (which requires autoregressive decoding of up to 80 tokens) required approximately 6 hours for 41,300 prompt-model pairs, or approximately 523 milliseconds per pair—roughly 10x the cost of MIA scoring. MIA scoring therefore adds roughly 10% to the cost of generation.

Stages 2 and 3 are negligible by comparison: logistic inference on a six-feature vector and a median over seven values per stock-date pair. Total overhead is therefore dominated by scoring, at roughly 1.15x the cost of unfiltered generation, and scales linearly in models and prompts while parallelising across GPUs. In deployment the binding constraint is observability rather than cost. Most commercial APIs do not expose per-token log-probabilities, which rules out Min-K%, Min-K%++ and the reference method; the loss-and-zlib variant reaches AUC 0.865 (Section 5.1) and represents what is achievable under that constraint.

Scalability to production settings requires consideration of three dimensions. First, model count: CMMD requires a minimum of three models with at least two distinct cutoff periods; adding more models improves partitioning granularity but increases MIA scoring cost linearly. In practice, five to ten models provide a good balance. Second, stock universe: expanding from 50 to 500 stocks increases cost linearly but benefits from GPU parallelism. Third, frequency: our daily rebalancing generates ~5,900 prompts per model per cycle. MIA scores exhibit temporal persistence—a stock's contamination profile changes slowly relative to model updates—so scores can be cached and refreshed weekly or monthly rather than recomputed daily, substantially reducing amortized cost.

For deployment with closed-source API models (GPT-4, Claude, Gemini), the pipeline adapts as follows. Per-token log-probabilities are not available from most API providers, precluding Min-K%, Min-K%++, and reference model methods. However, loss-based scoring can be approximated via the log-probability of the completion (available from some APIs), and zlib ratio can be computed from the prompt text alone. An API-compatible MCS variant using only loss and zlib features achieves reduced but still meaningful separation (estimated Cohen's d = 0.2–0.5 based on our two-feature ablation). Alternatively, the temporal proximity feature alone provides strong separation when cutoff dates are approximately known, as is the case for most major API providers who publish training data recency in their documentation.

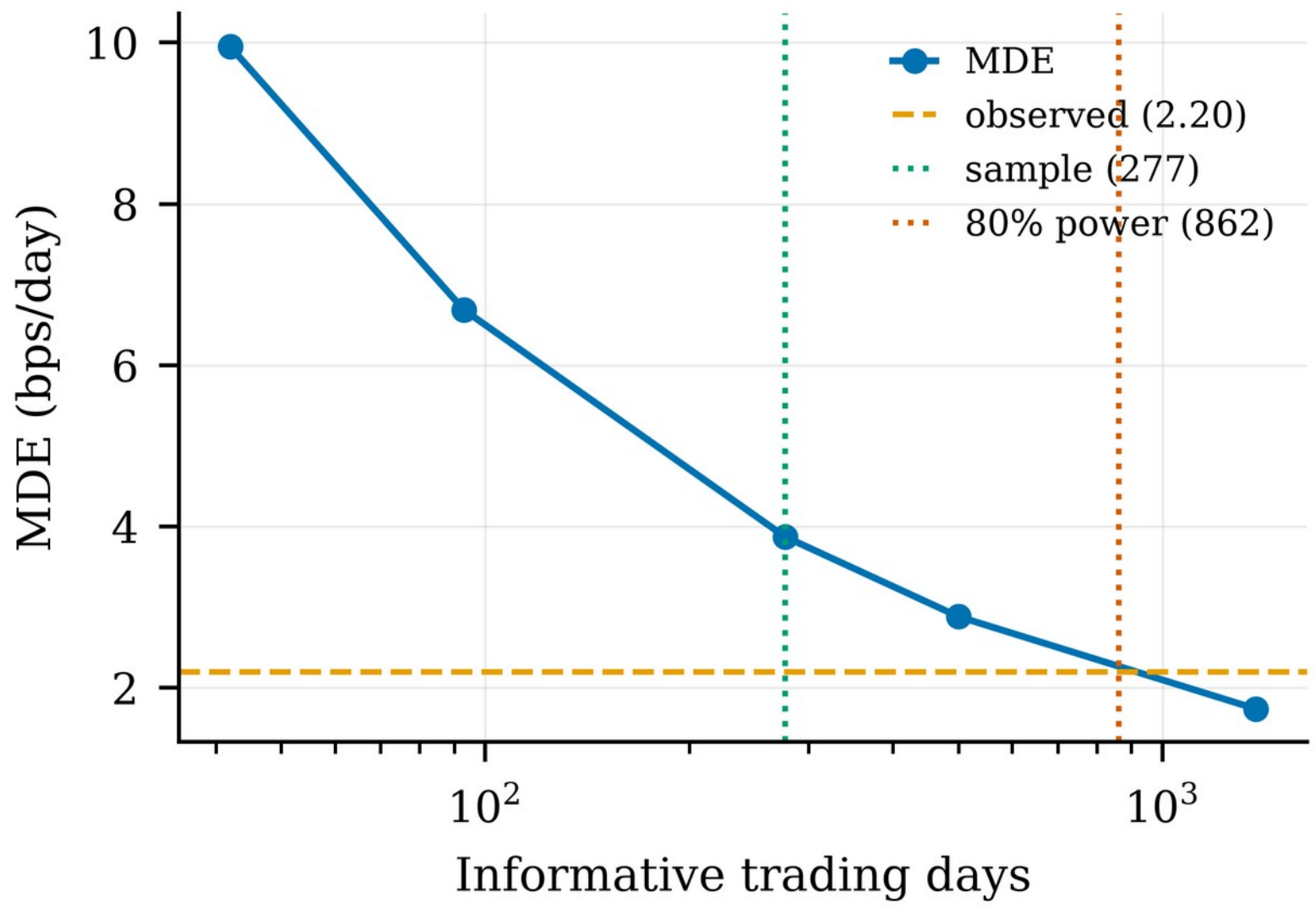

**Fig. 13.** Minimum detectable effect against the number of informative trading days, at 80% power and α = 0.05 two-sided, for the paired daily return difference between the filtered and unfiltered ensembles. The observed effect of 2.20 bps/day lies below the detectable threshold at the 277 dates available and would require roughly 862.

### *6.6 Limitations*

Several limitations qualify our findings. First, and most consequentially, the design has less power than we previously claimed. The submitted version put the requirement at roughly 250 trading days, derived from a Sharpe difference of 1.35 that was itself an artifact of the cost error. Recomputed on the corrected series, the paired daily difference between filtered and unfiltered ensembles has a standard deviation of 32.5 bps against a mean of 2.2 bps, so detecting an effect of that size at 80% power and α = 0.05 two-sided would require roughly 862 informative trading days. We have 277. Even the full daily series over this period, 1,375 days, would leave the design short, so we report a bounded null rather than treat the question as merely underpowered: our interval excludes Sharpe improvements above approximately +0.27. Second, the model lineup reaches 7B parameters and frontier models may behave differently. Third, alpha signals come from structured prompting rather than fine-tuned models. Fourth, MCS is fitted on the evaluation labels. We previously argued this mattered little because the dominant feature is deterministic; that argument was wrong in an instructive way, since the deterministic feature is precisely the one that recovers the label by construction. All detection results reported here are therefore cross-validated with folds grouped by calendar quarter. Fifth, the evaluation window is predominantly bullish, which may favour strategies preserving bullish signals; the 2022 sub-period, now 50 dates rather than the 17 available in the submitted version, shows CMMD at −2.97 bps/day against −3.69 for the unfiltered ensemble, a difference that is not significant.

## 7. Threats to Validity

### *7.1 Internal Validity*

Training cutoff dates provided by model cards are typically an approximation. Nonetheless, a high degree of similarity in MIA separation patterns among seven different models utilizing three different cutoff dates lends strength to the argument that slight differences in training cutoff dates will have little effect on overall results. Additionally, it is possible that the single reference model (GPT-2 base) used as the basis for comparison for the reference MIA method could result in biased comparisons, due to the possibility that GPT-2's familiarity with certain types of financial text correlates with how well the target models can memorize those same types of text.

### *7.2 External Validity*

Results cover 50 large-cap US equities. Large-caps appear disproportionately in training data, so memorization effects may be stronger here than for small-cap, international, or alternative assets. The Financial PhraseBank consists of pre-2016 European financial news, which may differ in linguistic characteristics from US financial text.

### 7.3 Construct Validity

We define memorization as the difference in the MIA scores for temporally In-Sample (IS) and Out-Of-Sample (OOS) prompts. Some of the text in the IS conditions were not even in the training set. It is possible that some of the OOS patterns are similar to those found in the training set. However, since we saw a very large separation (all 34 valid combinations had KS $p < 0.01$), it appears this is a good proxy. That said, we cannot rule out misclassification near the cutoff itself, where a prompt dated days either side of a model card's stated boundary may be labelled incorrectly.

### 7.4 Statistical Conclusion Validity

The asymmetry between our two levels of analysis is stark and worth stating plainly. The membership-inference results rest on 299,600 prompt-model pairs and are estimated with high precision; the portfolio results rest on 277 dates, of which the effective number carrying information about the filtered-versus-unfiltered comparison is smaller still. Under the published median split, the two strategies are identical on 56.7% of stock-date pairs, so those observations carry no information about the difference at all. We therefore treat the detection findings as well determined and the portfolio findings as a bounded null, and we quantify that bound in Section 6.6 rather than describing the study as merely underpowered.

The submitted version reported a first-order autocorrelation of 0.08 for the CMMD return series and used it to argue that Sharpe ratios were not materially inflated. That figure does not reproduce; the measured value is −0.003, and the +0.069 belongs to the Debiased Alpha series. The argument was in any case not applicable, since consecutive observations in our design are roughly 21 calendar days apart and cannot exhibit daily serial correlation. We withdraw it. All interval estimates reported here use a stationary block bootstrap with block length five, which is robust to dependence at that scale without requiring an assumption about its form.

The use of Yahoo Finance for price information also provides the opportunity for potential survivorship bias. The fifty S&P 100 constituents used in this study were chosen based on current membership. Therefore, the companies which were part of the index at some time previously but have since been deleted (for example due to a merger or acquisition, or if their market capitalization had decreased to the extent that they could no longer be included) are excluded from the analysis. In general, survivorship bias inflates backtest returns as the firms that survive are, on average, more successful than those that are removed from the index. While there will likely always be some degree of survivorship bias associated with any backtested strategy, it affects every strategy in Table 4 equally, so it does not advantage any filtering variant over the baselines. It does, however, inflate the equal-weight benchmark, which is relevant to the factor regressions in Section 5.2 where that benchmark is the only strategy attaining significant alpha.

## 8. Conclusion

We set out to build a signal-level filter for memorization-contaminated LLM forecasts, and the main value of this paper lies in the reasons it does not work. Three findings stand. Where in-sample status is defined by a training cutoff, any feature monotone in that cutoff recovers the label exactly, so composite scores mixing such features with membership-inference statistics report separation that is arithmetic rather than detection. The membership-inference statistics themselves discriminate between models rather than between memorized and unmemorized prompts (Fig. 5): raw scores reach AUC 0.99 at a fixed date, but three independent within-model normalizations bring this to 0.487–0.537, five of seven models fall below chance under within-model cross-validation, and the one apparent exception disappears when the comparison is restricted to windows around its own cutoff. And once transaction costs are applied symmetrically, no filtering variant beats the unfiltered ensemble, none attains significant five-factor alpha, and excluding the single weakest model outperforms every contamination-based filter we test.

Against the research questions: MIA scores separate in-sample from out-of-sample prompts, but the separation is largely attributable to differences between models rather than to memorization (RQ1); filtering does not improve risk-adjusted performance once costs are applied symmetrically, and a trivial model-exclusion baseline does better (RQ2); memorization detectability varies with recency, though recency is confounded with architecture and corpus and we no longer claim it dominates scale (RQ3); and threshold filtering is unstable across the percentile range, but the threshold-free alternative does not resolve that instability (RQ4). We make no claim about market efficiency on the basis of these results. What we offer instead is a set of failure modes, each specific enough to be checked against our released artifacts, for a class of method the field is adopting quickly and evaluating in ways that make it look better than it is. We note finally that He et al. [9], approaching from the prevention side, find the performance gap between chronologically consistent and conventionally trained models to be modest in a comparable asset-pricing task. Their result bounds how much any filtering method could recover; ours indicates that the membership-inference signal such a method would rely on is not reliable enough to recover it.

Several directions follow from a negative result of this kind. The most direct is to ask whether membership inference can be made to work in this setting at all, rather than assuming it can: an evaluation that varies memorization status while holding model identity fixed, for instance by comparing checkpoints of a single model trained to different cutoffs, would isolate the quantity our within-date test cannot. Second, the saturation and label-definition failures we document are properties of the evaluation design rather than of our particular models, and are worth checking in the several concurrent papers that key in-sample status to published cutoff dates. Third, extending to frontier models via API is possible in principle but constrained by observability, since per-token log-probabilities are generally unavailable; our two-feature variant reaches AUC 0.865 and indicates what is achievable under that constraint. Fourth, if black-box statistics cannot separate memorization from model identity, white-box methods may: sparse autoencoder analysis or causal circuit tracing could establish whether memorized financial facts occupy identifiable features, which would sidestep the confound we document rather than work around it. Fifth, a longer daily series across additional asset classes would reduce the minimum detectable effect, though Section 6.6 indicates that even the full daily series over this window would remain underpowered for the effect size observed.

## 9. Data Availability Statement

The financial price data used in this study were obtained from Yahoo Finance (https://finance.yahoo.com) and are publicly available. The Financial PhraseBank corpus [25] is publicly available at https://huggingface.co/datasets/financial_phrasebank. All language models evaluated in this study are publicly available open-weight models accessible through HuggingFace: GPT-2 (124M, 355M, 774M), TinyLlama 1.1B Chat, Phi-2 (2.7B), Qwen2.5-3B Instruct, and Qwen2.5-7B Instruct. The prompt templates and experimental configuration details are provided in Sections 3 and 4 of this paper. All artifacts required to reproduce every number in this paper are released publicly: MIA scores, alpha signals with untruncated model completions, portfolio return series, per-model generation checkpoints, the analysis notebooks, and an audit script that regenerates each reported quantity from the released data. The repository also contains a corrections file documenting the discrepancies between this revision and the submitted version, with the affected quantity, the corrected value, and the computation that produces it.

# CRediT Author Statement

**Dip Roy**: Conceptualization, Methodology, Software, Formal Analysis, Investigation, Writing – Original Draft, Visualization. **Rajiv Misra:** Supervision, Writing – Review & Editing. **S. K. Singh:** Writing – Review & Editing. **Anisha Roy:** Data Curation, Validation, Writing – Review & Editing